\documentclass[journal]{IEEEtran}
\IEEEoverridecommandlockouts
\usepackage[lmargin=18mm, rmargin=18mm,tmargin=18mm,bmargin=18mm]{geometry}
\usepackage{algorithm}
\usepackage{algorithmic}
\usepackage[T1]{fontenc}
\usepackage{mathrsfs}
\usepackage{multirow}
\usepackage{color}
\usepackage{cite}
\usepackage{verbatim}
\usepackage{amsmath}
\usepackage{amsthm}
\usepackage{array}
\usepackage{graphicx}
\usepackage{subfig}
\usepackage{makecell}
\usepackage{verbatim}
\usepackage{bbding}
\usepackage{pifont}
\usepackage{wasysym}
\usepackage{amssymb}
\usepackage{booktabs}
\usepackage{dutchcal}
\usepackage[normalem]{ulem}
\usepackage[justification=centering]{caption}
\usepackage{xcolor}
\usepackage[normalem]{ulem} 
\newcommand\hl{\bgroup\markoverwith
  {\textcolor{yellow}{\rule[-.5ex]{2pt}{2.5ex}}}\ULon}
\def\BibTeX{{\rm B\kern-.05em{\sc i\kern-.025em b}\kern-.08em
    T\kern-.1667em\lower.7ex\hbox{E}\kern-.125emX}}
\begin{document}

\title{A Lightweight Transmission Parameter Selection Scheme Using Reinforcement Learning for LoRaWAN}


\author{
\IEEEauthorblockN{Aohan Li,~\IEEEmembership{Member,~IEEE}, Ikumi Urabe, Minoru Fujisawa,\\ So Hasegawa,
Hiroyuki Yasuda, Song-Ju Kim and Mikio Hasegawa,~\IEEEmembership{Member,~IEEE}}
\thanks{Aohan Li is with the Graduate School of Informatics and Engineering, The University of Electro-Communications, Tokyo, Japan 
(Email: aohanli@ieee.org)}
\thanks{Ikumi Urabe, Minoru Fujisawa, and Mikio Hasegawa are with the Department of Electrical Engineering, Tokyo University of Science, Tokyo, Japan (Email:\{i-urabe,m-fujisawa\}@haselab.ee.kagu.tus.ac.jp, hasegawa@ee.kagu.tus.ac.jp)}
\thanks{So Hasegawa is with the
Social ICT System Laboratory, National Institute
of Information and Communication Technology
(NICT), Japan,
and
the Department of Electrical Engineering, Tokyo University of Science, Tokyo, Japan (Email: so-hasegawa@nict.go.jp)}
\thanks{Hiroyuki Yasuda is with the Graduate School of Law and Politics, The University of Tokyo, Tokyo, Japan. (Email: yasuda@g.ecc.u-tokyo.ac.jp)}
\thanks{Song-Ju Kim is with the Department of Electrical Engineering, Tokyo University of Science, Tokyo, Japan, and the SOBIN Institute LLC, Kawanishi, Japan. (Email: kim@sobin.org)}
}

\maketitle

\begin{abstract}
The number of IoT devices is predicted to reach 125 billion by 2023.
The growth of IoT devices will intensify the collisions between devices, degrading communication performance.
Selecting appropriate transmission parameters, such as channel and spreading factor (SF), can effectively reduce the collisions between long-range (LoRa) devices. However, most of the schemes proposed in the current literature are not easy to implement on an IoT device with limited computational complexity and memory. To solve this issue, we propose a lightweight transmission-parameter selection scheme, i.e., a joint channel and SF selection scheme using reinforcement learning for low-power wide area networking (LoRaWAN). In the proposed scheme, appropriate transmission parameters can be selected by simple four arithmetic operations using only Acknowledge (ACK) information. 
Additionally, we theoretically analyze the computational complexity and memory requirement of our proposed scheme, which verified that our proposed scheme could select transmission parameters with extremely low computational complexity and memory requirement.
Moreover, a large number of experiments were implemented on the LoRa devices in the real world to evaluate the effectiveness of our proposed scheme. The experimental results demonstrate the following main phenomena. (1) Compared to other lightweight transmission-parameter selection schemes, collisions between LoRa devices can be efficiently avoided by our proposed scheme in LoRaWAN irrespective of changes in the available channels.
(2) The frame success rate (FSR) can be improved by selecting access channels and using SFs as opposed to only selecting access channels. (3) Since interference exists between adjacent channels, FSR and fairness can be improved by increasing the interval of adjacent available channels.
\end{abstract}

\begin{IEEEkeywords}
LoRaWAN, Lightweight Transmission Parameters Selection Scheme, Reinforcement Learning.
\end{IEEEkeywords}

\section{Introduction}
\label{sect:introduction}
Internet of Things (IoT) plays an important role in modern industrial and smart cities. In recent years, IoT devices have been increasingly applied in various emergent applications, such as smart transportation, e-health, real-time control systems, and smart energy \cite{5}. By 2023, it is predicted that IoT devices will reach 125 billion, while IoT data generated by IoT devices is predicted to be reach 79.4 zettabytes (ZB) \cite{1}. IoT networks consist of different IoT devices using various protocols, such as ZigBee, low power wide area network (LPWAN), 5G network, and wireless local area network (WLAN) \cite{2}. Among these protocols, LPWAN is regarded as an up-and-coming technology due to its low energy consumption over long-distance and support for many IoT devices to access the network simultaneously. The most representative LPWAN technologies include Sigfox\footnote{Sigfox-The Global Communications Service Provider for the Internet of Things (IoT). https://www.sigfox.com/en}, low-power wide area networking (LoRaWAN), NB-IoT\footnote{A. Lombardo, S. Parrino, G. Peruzzi, and A. Pozzobon, "LoRaWAN Versus NB-IoT: Transmission Performance Analysis Within Critical Environments," \emph{IEEE Internet Things J.}, vol. 9, no. 2, pp. 1068-1081, May 2021.}, and LTE Cat. In this paper, we focus on LoRaWAN technology that has been widely implemented on several IoT devices \cite{4,6}.
Most LoRaWAN devices use a simple mechanism, such as pure ALOHA to access the channel in order to ensure low power consumption \cite{7}. In pure ALOHA, when the long range (LoRa) device needs to transmit data, it will send it at anytime. If a collision occurs during transmission, leading to a transmission fail, the LoRa device will wait a random amount of time and then re-transmit the data. Since pure ALOHA does not perform channel sensing before transmission, the network performance will drop sharply as the amount of data increases. 

To solve this problem, researchers and developers introduced cognitive radio (CR) technology into LoRaWAN, using which LoRa devices can dynamically use spectrum resources to avoid channel collisions and improve spectrum efficiency \cite{8, 9}.
Apart from access channels, spreading factor (SF) is another important factor that affects the performance of LoRaWAN \cite{22}. 
Chirp spread spectrum (CSS) technology is adopted by the physical layer of LoRa protocol, where chirp refers to the signals whose frequency varies linearly with time in the available bandwidth. Multiple data bits can be modulated in one chirp. The value of SF determines the chirp rate. For instance, the SF is 9 when 9 data bits are modulated in a chirp. The values of SF range from 7 to 12. A lower SF can shorten the on-air time of the packet but reduce its ability to resist noise. Conversely, higher SF can improve the ability to resist noise but on-air time will be longer\cite{23}.
The relationship between time on air, bandwidth (BW), and SF is given in Table I.
Since the access channel and SF directly affect the collisions among LoRa devices, 
it is extremely important to adjust them to achieve high communication performance.

\begin{table}[]
   \caption{Time on air (ms) when the payload is 50 bytes}
    \centering
    \begin{tabular}{|c|c|c|c|c|c|c|c|}

\hline
\multicolumn{1}{|c|}{\multirow{2}*{BW}}&\multicolumn{6}{|c|}{{SF}}
\\
\cline{2-7}
\multicolumn{1}{|c|}{}&7&8&9&10&11&12
\\
\toprule[1pt]
62.5&308&543&903&1642&2957&5587
\\
\hline
125&154&267&452&821&1479&2793
\\
\hline
250&77&133&226&411&739&1397
\\
\hline
500&38&67&113&205&370&698
\\
\hline
    \end{tabular}
    \label{bs2}
\end{table}

In recent years, many researchers have studied the automatic channel and SF adjustment schemes for LoRaWAN to achieve a good communication performance. The current literature in this area can be broadly divided into two categories: centralized schemes \cite{29, 30, 31, 32, 25, 10, 33, 34, 35, 36, 7, 38, 28, 37, 24, 11, 26, 27}, and distributed schemes \cite{39,12,42,43,44}.
Most of the autonomous communication parameters adjustment schemes are centralized where the parameters adjustment is aided by Gateway (GW). However, almost all of the centralized schemes face the following three disadvantages. 
First, prior information, such as location, packet, channel, and event information, are required for GW to select transmission parameters. 
Second, the GW needs to inform the LoRa devices about their used transmission parameters, which increases the transmission delay and spectrum resource consumption. Third, it is necessary to listen to GW in real time to obtain the transmission parameters for LoRa devices, which results in additional energy consumption. As described previously, the energy and spectrum consumption of the centralized schemes make it impossible to be implemented in future massive IoT networks  \cite{12}.

Conversely, in the distributed schemes, the IoT device itself selects the transmission parameters without the aid of GW. Therefore, spectrum and energy consumption, and communication latency can be reduced, which makes it a good candidate for future implementations in LoRaWAN. Literature on fully distributed transmission parameters selection schemes is pretty limited \cite{39,12,42,43,44}.
In \cite{39} and \cite{12}, although decentralized transmission parameters selection schemes are proposed for LoRaWAN, the implementation of the schemes in practical LoRaWAN is not considered. In our previous work \cite{42}, and references \cite{43} and \cite{44}, the proposed schemes are implemented on practical LoRaWAN systems. However, no specific LoRaWAN characteristic was considered in the problem formulation, such as SF selection.


To further improve the communication performance of LoRaWaN while considering the characteristics of LoRa devices, we propose a tug-of-War (ToW) dynamics \cite{13,14,15,19}-based decentralized transmission parameters selection scheme for LoRaWAN. 
In the proposed scheme, the channel and SF are selected by LoRa devices.
The ToW dynamics is a unique parallel search multi-armed bandit (MAB) algorithm, which aims to maximize the sum of rewards for a dynamic environment \cite{14, 19}.
Since the ToW dynamics can make decisions using only simple four arithmetic operations, it has been applied to solve channel selections in IoT networks with limited computation, and memory ability \cite{15,16,17}.
The following are the main contributions of this paper:


\begin{itemize}
\item 
A lightweight ToW dynamics-based joint channel and SF decision scheme is proposed for LoRaWAN.
In the proposed scheme, the LoRa device can select its access channel and SF based only on Acknowledgement (ACK) information.

\item 
We theoretically analyze the memory requirement and computational complexity of our proposed scheme, which shows that our proposed scheme can select channel and SF with low computational complexity and memory.
\item 
We implement the proposed scheme on practical LoRa devices and evaluate the performance using frame success rate (FSR). In addition, we compare the performance of the proposed scheme to other state-of-the-art MAB-based lightweight transmission parameters selection schemes. From the experimental results, we observe that the proposed scheme can obtain higher FSR than the other schemes in the LoRaWAN system irrespective of changes in channel availability. 
Additionally, we also show that, when the available channels are dynamically varied, LoRa devices can switch to available channels more quickly using our scheme as compared to other schemes in the scenario.

\item
To evaluate the effectiveness of the proposed scheme in a more realistic communication environment, we evaluate our proposed scheme in LoRaWAN coexisting with other IoT networks, i.e., Wi-SUN networks. Wi-SUN is also a type of IoT device that functions at the same 920 MHz frequency band as LoRa devices in Japan. 
The experimental results show that higher FSR can be achieved using the proposed scheme as compared to other schemes. Hence, LoRa devices using our proposed scheme can efficiently avoid collisions from other LoRa devices and IoT devices using the other protocol.
\item
Interference between adjacent channels is also considered in this paper. By evaluating the performance using fairness, we can observe that FSR and fairness can be improved by increasing the interval between adjacent channels, which verified that the interference in the practical IoT systems is not only from other IoT devices but also from adjacent channels. 
\end{itemize}

The remainder of the paper is organized as follows. Related work is reviewed in Section \uppercase\expandafter{\romannumeral2}. The system model and problem formulation are introduced in Section \uppercase\expandafter{\romannumeral3}. 
Section \uppercase\expandafter{\romannumeral4} presents the proposed transmission parameters selection scheme.
Experimental results are demonstrated in Section \uppercase\expandafter{\romannumeral5}.  Finally, the paper is concluded in Section \uppercase\expandafter{\romannumeral6}.

\section{Related Work}
\label{sect:introduction}
Related literature on transmission parameter selection schemes for LoRaWAN is introduced in this section.
We first review centralized schemes, followed by decentralized schemes.

\subsection{Centralized Resource Allocation}
Centralized schemes can be divided into three categories: optimization schemes \cite{29, 30, 31, 32, 25, 10, 33}, machine learning schemes \cite{34, 35, 36, 7, 38}, and other schemes \cite{28, 37, 24, 11, 26, 27}. 
In the rest of this section, each category of related work will be introduced sequentially: 

\subsubsection{Optimization schemes} In the optimization schemes \cite{29, 30, 31, 32, 25, 10, 33}, the optimization problems related to resource allocation are formulated first. Then, optimization solvers are used to solve the formulated problems with a single variable \cite{32, 33}. For the optimization problems with multiple variables, the problem is generally split into sub-optimization problems, and each sub-optimization problem is solved respectively \cite{29, 30, 31, 25, 10}.

In \cite{32}, three optimization problems were proposed, which were termed OPT-MAX, OPT-DELTA, and OPT-TP. The OPT-MAX and OPT-DELTA consider the minimization problem of maximum probability of intra-SF collisions and intra-SF and inter-SF collisions, respectively. The OPT-TP was used to minimize the total energy consumption of all LoRa devices by allocating appropriate power. In this study, IBM ILOG CPLEX was adopted to solve the formulated optimization problems.
The maximization problem of the average system packet success probability was formulated in Ref. \cite{33}. To achieve sub-optimal solutions to the formulated maximization problem, an SF allocation scheme using a global optimization solver for each traffic was proposed.
In \cite{29}, a combinatorial optimization problem was formulated that simultaneously satisfied the following optimizations. (1) The maximization problem of the minimum average transmission rate of the uplink communication for LoRa devices. (2) The optimization problem of the allocations for energy harvesting (EH) time, power, and SF. In this work, the SF was first assigned based on received signal strength indicator. Then, the EH time was optimized using the one-dimensional exhaustive search method. Finally, the transmit power was optimized using the concave-convex procedure for the given EH time and SF.
In \cite{30}, an optimization problem was formulated to maximize the minimum throughput for all LoRa users. Joint optimization of the allocations of the SF, duty cycle, and transmit power were considered to achieve optimal solutions. To achieve that, an iterative balancing method was proposed to adjust the SF, duty cycle, and transmit power based on spatial distributions of LoRa users and average channel statistics.
In \cite{31}, the minimum throughput of LoRa devices in uplink communications was considered. An optimization problem was formulated to maximize the minimum throughput. The optimization problem was constrained by power, intra-SF, and inter-SF with adjustable variable power and SF. The formulated problem was a mixed-integer non-linear optimization. To solve this problem, two sub-optimization problems were introduced, i.e., SF allocation and joint power and SF allocation. To solve the first sub-problem, the many-to-one matching method was adopted to match SFs and LoRa devices. Next, the second sub-problem was solved by assigning power under fixed SF and SF under fixed power.
In \cite{25}, a joint SF and power allocation and user scheduling problem to maximize the energy efficiency was formulated. To solve the formulated problem, the authors proposed a matching theory-based scheduling, a distance-based SF allocation, and a sequential convex and generalized fractional programming-based iterative power allocation scheme.
In \cite{10}, an optimization problem to maximize the network throughput with the constraints of dynamic energy harvesting was formulated. To solve the formulated optimization problem, a matching theory-based channel allocation scheme and a Markov decision process-based power allocation scheme were proposed to assign optimal channels and power to LoRa users, respectively.

Although near-optimal performances can be achieved by the optimization methods with low computational complexities as compared to exhaustive search,
it may face the following problems.
First, prior information is necessary for GW to assign transmission parameters (e.g., location of the LoRa users, length of the packet, traffic per unit time, channel information, and probability of an event). In addition, GW needs to provide the transmission parameters to LoRa devices, which may require extra spectrum resource computation and increase the communication latency. 
Second, real-time listening for LoRa devices is necessary to obtain the transmission parameters from GW, which may result in high energy consumption.
Third, since the joint optimization is solved, the correlation of physical layer (PHY) parameters maybe not well considered.
Fourth, transmission parameters need to be reallocated when prior information changes, which will increase the computation, communication resource, and energy consumption to a great extent.

\subsubsection{Machine learning-based schemes}
In machine learning-based schemes \cite{34, 35, 36, 7, 38}, learning method is run on the GW to learn the PHY parameter allocation for LoRa devices.
Machine learning-based methods can produce the best action policy by learning the statistical distributions of dynamic environmental parameters that can achieve the maximum reward.
In current literature, except for \cite{35} that only considered the SF allocation, the other works considers the PHY parameter correlation \cite{34, 36, 7, 38}.

In \cite{34}, a reinforcement learning method was developed taking into account the channel assignment and amount of energy that will be harvested to improve the energy efficiency of LoRaWAN.
In \cite{35}, an long short-term memory (LSTM) based SF assignment scheme was proposed for noncooperation networks with minimum synchronization. In the proposed scheme, the assignment of SF was based on the success rate of each SF predicted by the LSTM network. 
In \cite{36}, the optimization problem of maximizing the energy efficiency, quality of service (QoS), and reliability by intelligently configuring the parameters, such as SF and transmission power, was considered. A deep deterministic policy gradient (DDPG) was proposed to solve the formulated problem. Moreover, a multi-agent DDPG based on the transfer learning method was proposed to accelerate the training process.
In \cite{7} and \cite{38}, joint SF and power assignment schemes for LoRaWAN were proposed, where the SF and power are assigned using deep reinforcement learning (DRL), and Q-learning methods, respectively.

Although machine learning-based schemes consider the PHY parameter correlation, they still face the first and second drawbacks as optimization schemes. Additionally, these methods take a long time to train the model until convergence. Hence, when the communication environment changes, the model requires to be trained again, which may produce long communication latency.

\subsubsection{Other schemes} 
In addition to the optimization and machine learning methods, some other centralized methods exist, such as game theory-based methods, signal-to-noise-ratio (SNR)-based methods, distance-based methods \cite{28, 37, 24, 11, 26, 27}.

In \cite{28}, a joint optimization problem to maximize the energy efficiency and the ratio of packet delivery was considered. The optimal settings of SF and power and signal-to-interference-and-noise ratio (SINR) could be achieved by power allocation at the network server based on the proposed game-theoretic framework.
In \cite{37}, to reduce the interference between LoRa devices, an estimation method on the transmission time duration for suitable SFs using game theory was proposed.
In \cite{24}, SF was assigned using GW based on signal-to-noise ratio (SNR) to improve the scalability of LoRaWAN. It was assumed that SNR is known at the GW.  
In \cite{11}, two-step lightweight scheduling was presented. 
First, the allowed SFs and power for LoRa devices on each channel were specified dynamically. Next, each LoRa device decided its access channel, SF, and transmission power by itself.
In \cite{26}, SF was assigned based on channel conditions. 
Specifically, to improve the symbol error rate (SER) and fairness of the overall performance, the devices with poorer channel conditions were assigned higher SFs, while the devices with higher channel conditions were assigned lower SFs.
In \cite{27}, an SF and channel allocation policy was proposed. SF was allocated based on the distance between the LoRa device and GW, while the channel was allocated to minimize the average outage probability in the cell, which was solved by the exhaustive search solution. 

Although some of the schemes are easy to be achieved, they still face the same disadvantages as other centralized schemes, i.e., listening to the GW and prior information transmission between GW and LoRa devices are necessary, which increases the resource and energy consumption and communication latency.

\subsection{Decentralized Parameter Selection}
In decentralized transmission parameters selection schemes, each LoRa device selects its transmission parameters by itself. Hence, compared to centralized schemes, the spectrum and energy efficiency can be improved.
Existing literature on fully decentralized parameter selection is limited \cite{39, 12, 42, 43, 44}.
In \cite{39}, exponential weights for exploration and exploitation algorithm was proposed to autonomously decide the SFs for LoRa devices.
In \cite{12}, a low-complexity distributed learning scheme that were applicable in LoRaWAN was first studied. 
The analytical and simulation results reveal that the proposed MAB-based scheme could improve the data transmission success probability and the battery lifetime of LoRa devices. However, the implementation of the proposed scheme on practical LoRa devices with extremely limited computing ability and memory under dynamic LoRaWAN was not considered. 
Although references \cite{42, 43, 44} consider the implementation of the schemes on practical LoRaWAN, the specific characteristics of LoRaWAN was not considered in the problem formulation, such as SF selection.

\section{System Model and Problem Formulation}
\label{sect:systemmodel}

\begin{figure}
    \centering
    \includegraphics[width=80mm, height=55mm]{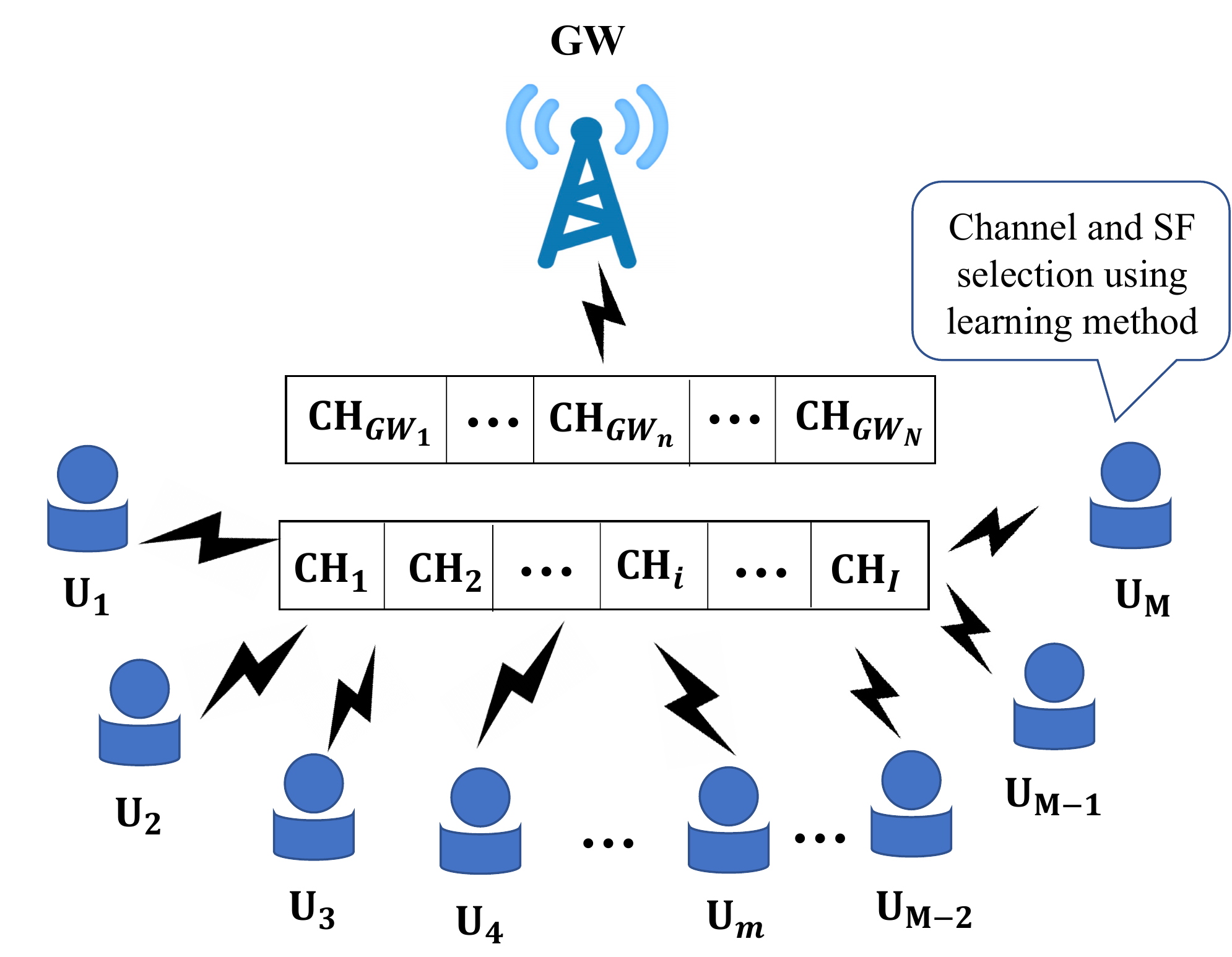}
   \caption{System model.}
    \label{fig:systemmodel}
\end{figure}

We consider an uplink LoRaWAN system, as shown in Fig. 1, where there is one multichannel GW with $N$ available channels and multiple LoRa devices with $I$ available channels. 
Denoting the set of LoRa devices as $\mathcal{U}=\{U_1, U_2,...,U_m,...,U_M\}$, where $M$ and $U_m$ denote the number of LoRa devices and the $m$-th LoRa device, respectively. 
Denote the available channel sets for GW and LoRa devices as $CH_{GW}=\{CH_{GW_1}, CH_{GW_2},...,CH_{GW_n},...,CH_{GW_N}\}$ and $CH_{LD}=\{CH_1, CH_2,...,CH_i,...,CH_I\}$, respectively, where $CH_{GW_n}$ and $CH_i$ denote the $n$-th  available channel and the $i$-th available channel for GW and LoRa devices, respectively. Assume that the bandwidth for each channel is B. 
Denote the SF set for the LoRa devices as $SF=\{SF_1, SF_2,...,SF_s,...,SF_S\}$, where $SF_s$ denotes the $s$-th SF and $S$ is the number of SF values, which equals to 7. 
The values of the elements in SF set are {6, 7, 8, 9,10, 11,12}.


The lightweight decentralized reinforcement learning scheme is loaded on the LoRa devices. At each decision time, the LoRa device selects its access channel and SF based on the reinforcement learning scheme. Denote $TI$ as the transmission interval between two times of decision. That is, the frequency of a LoRa device make a channel and SF selection decision is once per $TI$ times. After each decision, the LoRa device performs clear channel assessment (CCA) to check whether the availability of the selected channel. If the channel is available, then the LoRa device sends message with payload $L$ to GW using that channel. If the transmission is successful, ACK information will be received by the LoRa device. Otherwise, the LoRa device will receive negative acknowledgment (NACK) information. We assume that each device can re-transmit $TR$ times after each decision. 
If the ACK information can be received by the LoRa device within the TR times re-transmission, we consider the transmission as to be successful. Otherwise, we consider the transmission as a failure.

Denote $r_m^{is}(t)$, which indicates if the transmission is successful for LoRa device $U_m$ using channel $CH_i$ and $SF_s$ at the $t$-th decision. Its expression is given as follows:

\begin{equation}
    r_m^{is}(t) =
    \begin{cases}
    1, \text{Successful transmission}. \\
    0, \text{otherwise}.
    \end{cases}
    \label{eq:n_ik}
\end{equation}
Denote $n_m^{is}(t)$, which indicates if LoRa device $U_m$ attempts to transmit messages using channel $CH_i$ with $SF_s$ at the $t$-th decision, and it is given as follows:
\begin{equation}
    n_m^{is}(t) =
    \begin{cases}
    1,\text{transmission attempted}
    \\
    0,\text{otherwise}.
    \end{cases}
    \label{eq:n_ik}
\end{equation}
At the $t$-th time decision, the FSR of the LoRaWAN system can be given by

\begin{equation}
    P(t) = \sum_{m=1}^{M}\sum_{i=1}^{I}\frac{r_m^{is}(t)}{n_m^{is}(t)},
\end{equation}
which provides the ratio of the successful transmission times and the attempted transmission times until the $t$-th decision for the LoRaWAN system.


This paper aims to maximize the FSR of the LoRaWAN system by intelligently selecting access channels and SF. The objective function is given as follows:
\begin{equation}
     \max_{\{i, s\}}\sum_{t=1}^{T}P(t),
\end{equation} 
where $\{i, s\}$ denotes the channel-SF pair; that is, LoRa device $U_m$ transmits messages using channel $CH_i$ and SF $SF_s$.


\section{TOW Dynamics-Based Transmission Parameter Selection Scheme}
\label{sect:proposedmethod}
In this section, we present our proposed transmission parameter selection scheme based on a MAB algorithm called ToW dynamics. In the proposed scheme, the transmission parameter problem is first transformed into an MAB problem. Then, the ToW dynamics is used to solve the transformed MAB problem. In the remainder of this section, we first introduce the relevance  between the MAB problem and our formulated optimization problem. Then, we present our proposed scheme. Finally, we analyze the memory requirement and computational complexity of the proposed scheme.


\subsection{MAB Problem and Transmission Parameters Selection in LoRaWAN}
MAB is a representative statistical model in reinforcement learning, which can solve decision problems by balancing exploration and exploitation. In the MAB problem, each arm provides a random reward from an unknown probability distribution.
The MAB aims to maximize the sum of rewards earned by a sequence of lever pulls. It is shown that the uncertainty of dynamic spectrum access can be well tracked by MAB algorithms \cite{3}. Since MAB algorithms are simple, they can be implemented in a low computational complexity scenario. 
Our formulated problem aims to maximize the accumulated FSR of the LoRaWAN system by suitable choice of access channel and SF.
The maximization of the accumulated FSR can correspond to the maximization of rewards, while the selection of the appropriate access channel and SF can correspond to the sequence of lever pulls in the MAB problem. By setting each LoRa device as a machine while setting the channels and SF as the arms of the machine, the joint channel and SF selection problem in LoRaWAN can be formulated as a MAB problem. To make it easier to understand, Fig. 2 shows the relatedness between the MAB problem and the CH-SF selection in LoRaWAN. 

\begin{figure}
    \centering
    \includegraphics[width=90mm, height=40mm]{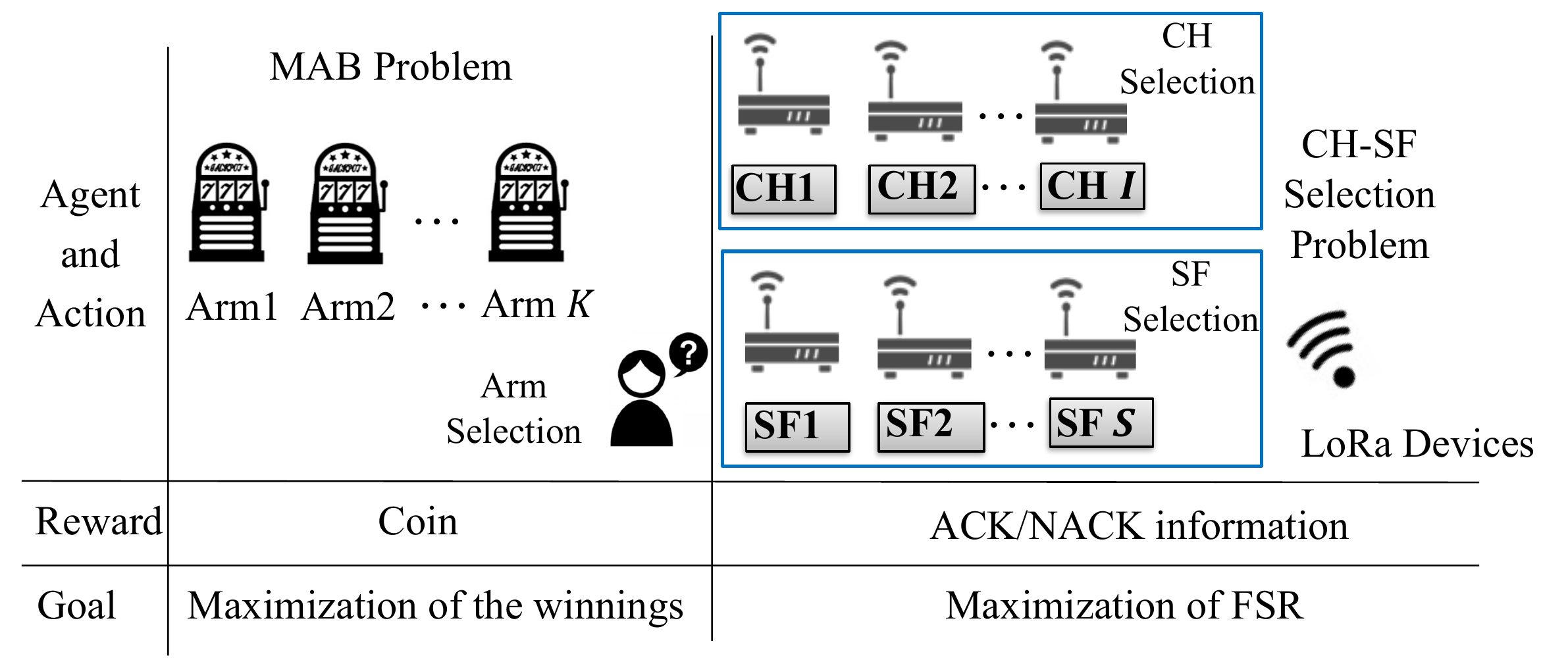}
   \caption{Comparison of the MAB and CH-SF selection problems.}
    \label{fig:systemmodel}
\end{figure}

\subsection{ToW Dynamics-Based CH-SF Selection}
In this subsection, we introduce our proposed ToW dynamics \cite{13,14,15,19}-based CH-SF selection approach. 
We evaluated the FSR for different structure designs of the CH-SF selection MAB problems beforehand, and the results show that independent arms for CHs and SFs can obtain the highest FSR. Hence, we introduce the independent CH-SF selection approach here. 

As shown in Fig. 2, the CH and SF selections are performed using two independent MAB sets. The number of arms in the two MAB sets are $I$ and $S$, respectively, which correspond to the number of channels and SFs in our system. Denote $j$ as the indicator to indicate the selection of CH or SF, where $j\in\{1,2\}$. We assume that $j=1$ and  $j=2$ correspond to the selection of CH and SF, respectively. 
In the proposed scheme, access CH $k_1$ and SF $k_2$ are selected based on the following policy by the LoRa device itself.
\begin{subequations}
\begin{align} 
 k_1^{\ast}= \mathrm{argmax}_{k_1 \in {CH_U}} X_{k_1}(t), \\
 k_2^{\ast}= \mathrm{argmax}_{k_2 \in {SF}} X_{k_2}(t),
\end{align}
\end{subequations}
where $k_1$ and $k_2$ are the corresponding arms of the CH and SF, respectively. 
At each time of decision, the LoRa device selects the CH and SF with maximum $X_{k_1}(t)$ and $X_{k_2}(t)$ as its access CH and using SF, respectively.
The LoRa device selects its access CH and SF randomly when there exist multiple channels with the same maximum $X_{k_j}(t)$.
 
Here, $X_{k_j}(t)$ can be expressed as

\begin{equation}
    X_{k_j}(t)=Q_{k_j}(t-1)-\frac{1}{I-1}\sum_{{k_j}^{'}\neq {k_j}}{Q_{{k_j}^{'}}(t-1)+osc_{k_j}(t)}.
    \label{eq:k-argmax}
\end{equation}
In \cite{13}, the influence of $osc_{k_j}(t)$ on the efficiency of decision-making is studied, where
$osc_{k_j}(t)$ is defined as
\begin{equation}
    osc_{k_j}(t) = Acos\left(\frac{2\pi (t+{k_j}-1)}{D}\right),
    \label{eq:Q_k}
\end{equation}
Here, $Q_{k_j}(t-1)$ 
denotes the Q value of the $k$-th channel when $j=1$, and the $k$-th SF when $j=2$ at the $(t-1)$-th decision-making. 
$D$ is the number of arms. Hence,
$D=I$ when $j=1$, while $D=S$ when $j=2$. $Q_{k_j}(t-1)$ in equation (6) can be calculated by

\begin{align}
    Q_{k_j}(t)=
     \begin{cases}
            \alpha Q_{k_j} (t-1)+\Delta Q_{k_j} (t),\;\;\;\text{if}\; {k_j}={k_j}^{\ast}, \\
\alpha Q_{k_j} (t-1),\;\;\;\;\;\;\;\;\;\;\;\;\;\;\;\text{otherwise},
        \end{cases}
    \label{eq:DeltaQ}
\end{align}
where $\alpha$ is defined as the discount factor. Its value range is $0\leq \alpha \leq 1$. The role of $\alpha$ is to make the scheme adapt more effectively to the dynamically changing communication environment by controlling the impact of past rewards 
when estimating the reward.
$\Delta Q_{k_j}(t)$ can be expressed as

\begin{align}
    \Delta Q_{k_j}(t)=
        \begin{cases}
            +1,\;\;\;\text{transmission is successful.} \\
-\omega_j(t),\;\;\;\text{otherwise}.
        \end{cases}
    \label{eq:DeltaQ}
\end{align}
In other words, if the transmission is successful using the selected channel and SF, the selected CH $k_1$ and the used SF $k_2$ obtain a reward, and the Q value corresponding to the selected CH $k_1$ and SF $k_2$ , i.e., $Q_{k_1}(t)$ and $Q_{k_2}(t)$, are updated by adding 1. Otherwise,
$Q_{k_j}(t)$ is updated by adding a penalty of $-\omega_j(t)$,
where $\omega_j(t) $ is derived by
\begin{equation}
    \omega_j(t) = \frac{p_{j_{\mathrm{1st}}}(t)+p_{\mathrm{j_{2nd}}}(t)}{2-p_{j_{\mathrm{1st}}}(t)-p_{j_{\mathrm{2nd}}}(t)},
    \label{eq:omega_i}
\end{equation}
where $p_{j_{\mathrm{1st}}}(t)$ and $p_{j_{\mathrm{2nd}}}(t)$ are the reward probabilities of the arm with the highest and second-highest reward probabilities at time $t$.
The reward probability $p_{k_j}(t)$ corresponding to CH $k_1$ and SF $k_2$ can be derived by

\begin{equation}
    p_{k_j}(t)=\frac{R_{k_j}(t)}{N_{k_j}(t)},
    \label{eq:p_ik}
\end{equation}
where $R_{k_j}(t)$ is the number of successful transmissions by selecting the slot machine $k_j$ until time $t$. $N_{k_j}(t)$ is the times to select the slot machine $k_j$ until time $t$.
$N_{k_j}(t)$ and $R_{k_j}(t)$ are given by:
\begin{equation}
    N_{k_j}(t) =
    \begin{cases}
    1+\beta N_{k_j}(t-1),\;\;k_j=k_j^{\ast}. \\
    \beta N_{k_j}(t-1),\;\;\text{otherwise.}
    \end{cases}
    \label{eq:n_ik}
\end{equation}
 
\begin{equation}
    R_{k_j}(t) =
    \begin{cases}
    1+\beta R_{k_j} (t-1),\;\;k_j=k_j^{\ast} \\\;\;\;\;\;\;\;\;\;\;\;\;\;\;\;\;\;\;\;\;\;\;\;\;\;\;\text{and transmission is successful.} \\
    \beta R_{k_j}(t-1),\;\;\text{otherwise.}
    \end{cases}
    \label{eq:n_ik}
\end{equation}
Here, $\beta$ is a forgetting factor, and its role is also to make the algorithm adapt the dynamic environment. 
It is used to suitably forget previous $N_{k_j}(t)$ and $R_{k_j}(t)$ to achieve the goal. 
Specifically, the fluctuation of $\omega_j(t)$ becomes smaller as the number of trials increases.
During a long-term operation, with an decrease in $\omega_j(t)$, 
%
there is a possibility that learning cannot follow the fluctuating environment. 
To prevent this problem, $\beta$ is introduced.


As summarized in Algorithm 1, the process of the proposed scheme is presented as follows.
First, we set the initial value of  
$Q_{k_j}(0)$, $R_{k_j} (0)$, $N_{k_j} (0)$, and $t$ to zero. Each LoRa device selects its access channel and SF randomly at the initial decision, i.e., when $t=0$.
When $t\ge 1$, the CH and SF will be selected based on equation (5).
After selecting the access channel and SF, the LoRa device sends a message using the selected channel and SF.
If the transmission is successful, the ACK information will be received at the LoRa device. Otherwise, the transmission is a failure. 
Then, the LoRa device updates the $\Delta Q_{k_j}(t)$ based on feedback from GW, i.e., whether to receive the ACK information, using equation (9).
According to equation (9), $\Delta Q_{k_j}(t)$ is increased by 1 if the transmission is successful. Otherwise, $\Delta Q_{k_j}(t)$ is decreased by $\omega(t)$ which is updated based on equations (10), (11), (12), and (13). Then, $Q_{k_j}(t)$ is updated using $\Delta Q_{k_j}(t)$ according to equation (8) followed by $X_{k_j}(t)$ being updated based on $Q_{k_j}(t)$ using equation (6). 
Until this point, only the $t$-th time decision process is completed.
Then, $t$ is increased by 1, i.e., $t = t + 1$. The process mentioned above will be repeated until $T$. 

\begin{algorithm}[ht]
  \caption{ToW dynamics-based CH-SF selection}
  \label{alg:towcs}
  \begin{algorithmic}[1]
    \REQUIRE $Q_{k_j}(0)$, $R_{k_j}(0)$, $N_{k_j}(0)$, $t=0$.
    \WHILE{$t\leq T$}
 \IF{t=0}  
     \STATE{Access channel and used SF are selected randomly.}
 \ELSE
    \STATE{Access channel and used SF are selected using equation (5).}
\ENDIF
 \IF{transmission is successful,}
 \STATE{Set $\Delta Q_{k_j}(t)=+1$}     
\ELSE
     \STATE{Updated
     $N_{k_j}(t)$ and $R_{k_j}(t)$ using equations (12) and (13).}
      \STATE{Calculate $p_{k_j}(t)$ based on $N_{k_j}(t)$ and $R_{k_j}(t)$ using equation (11).}
    \STATE{Calculate $\omega_j(t)$ based on $p_{k_j}(t)$ using equation (10).}
     \STATE{Set $\Delta Q_{k_j}(t)=-\omega_j(t)$} 
   \ENDIF
      \STATE{ $osc_{k_j}(t+1)$ and $Q_{k_j}(t)$ are updated using equations (7) and (8).}
    \STATE{ $X_{k_j}(t+1)$ is updated based on $osc_{k_j}(t+1)$ and $Q_{k_j}(t)$ using equation (6).}
    \STATE{$t = t + 1$.}
    \ENDWHILE
  \end{algorithmic}
\end{algorithm}

The operation process of the proposed scheme on LoRa devices are summarized as follows, while the flowchart is shown in Fig. 3. 
1) The access channel and SF are determined using the proposed CH-SF selection scheme by the LoRa device.
2) The LoRa device carries out CCA to determine whether the selected channel is available. 
3) If the selected channel is available, the message with payload $L$ is sent by the LoRa device using that channel with the selected SF.
4) If the LoRa device can receive the ACK information after sending the message, the transmission is successful. Otherwise, the message will be resent. 
If the ACK information can be received within $TR$ times re-transmission, the transmission can also be regarded as successful. Otherwise, the transmission is a failure.
5) The corresponding values related to the proposed schemes are updated according to equations (2)-(8).
6) LoRa devices turns to sleep model for $TI$ s.
7) The process from 1) to 6) will be repeated at transmission interval $TI$ until $T$.
In practical LoRaWAN systems, LoRa devices will keep operating until the battery is exhausted. That is, $T$ equals to the times of transmission attempts until the battery is exhausted. 

\begin{figure}
    \centering
    \includegraphics[width=80mm, height=120mm]{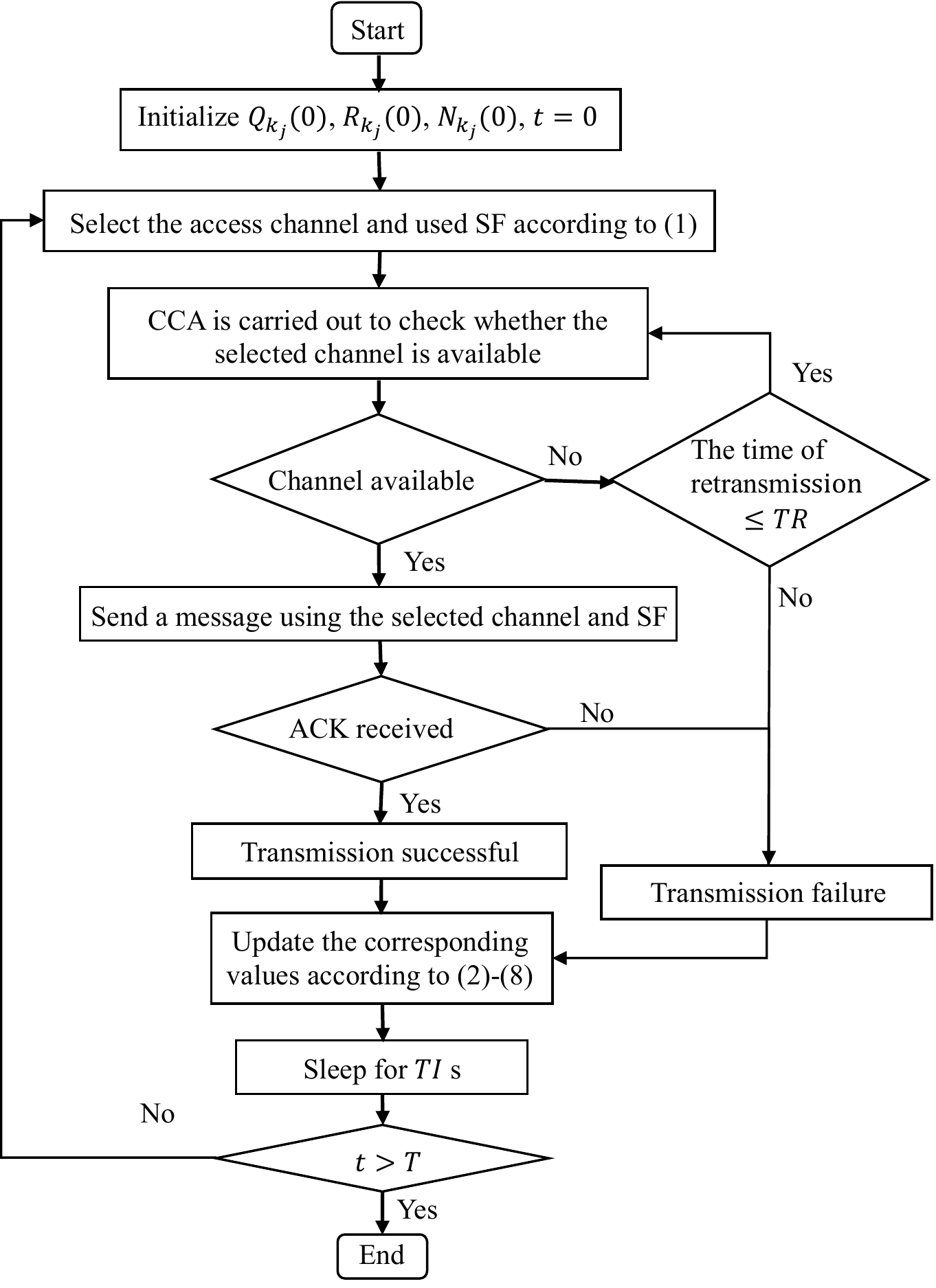}
   \caption{Flowchart of the ToW dynamics-based CH-SF selection.}
    \label{fig:systemmodel}
\end{figure}

\subsection{Theoretical Analysis of the Memory Requirement and Computational Complexity}
\subsubsection{Memory Requirement}
In our proposed ToW dynamics-based CH-SF selection scheme, only the values of $N_{kj}(t)$, $R_{kj}(t)$, and $Q_{kj}(t)$ for each arm are required to be stored.
Assume that the size of the value of the $N_{kj}(t)$, $R_{kj}(t)$, and $Q_{kj}(t)$ are $\mathcal{n}$bits, $\mathcal{r}$bits, and $\mathcal{q}$bits, respectively.
Since the number of arms is $I+S$ that equals to the summation of the number of IoT devices and SFs, the memory requirement for storing the values using our proposed scheme is only $(I+S)\times (\mathcal{n}+\mathcal{r}+\mathcal{q})$ bits.

\subsubsection{Computational Complexity} In our proposed ToW dynamics-based CH-SF selection scheme, each IoT device decides its access channel and SF by itself. Hence, the computational complexity for each time of iteration is only $O(1)$.


\section{Performance Evaluation}
\label{sect:experiment}
In this section, the performance of our proposed ToW dynamics-based CH-SF selection scheme is evaluated.
In addition, we compare it to several other lightweight schemes.
Specifically, the experimental settings of LoRaWAN built in this study is introduced first.
Then, the compared lightweight schemes are presented, i.e., UCB1-tuned based, $\epsilon$-greedy-based, and random transmission parameter selection schemes. 
We also evaluate the performance using FSR for the channel selection in our proposed scheme and compare it to other lightweight schemes with/without varying available channels. 
Moreover, 
the FSR of the CH-SF selection using our proposed scheme and other presented lightweight schemes are evaluated and compared under the built LoRaWAN with/without varying available channels.
Furthermore, considering practical IoT systems, we evaluate the FSR of CH-SF selection under the dynamical environment, where LoRa devices are coexisting with other types of IoT devices. 
In our experiment, Wi-SUN is used as the other type of IoT device, the details of which are discussed in a later section.
Finally, we investigate the interference between adjacent channels to improve the FSR and fairness. 

\subsection{Experimental Settings}

\begin{figure}
\centering
\subfloat[LoRaWAN system.]{
\label{fig:improved_subfig_a}
\begin{minipage}[t]{0.5\textwidth}
\centering
\includegraphics[width=70mm,height=40mm]{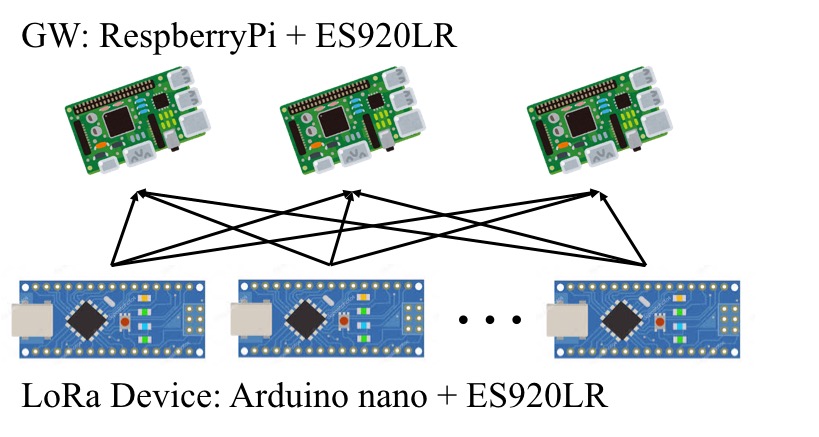}
\end{minipage}
}
\\ 
\subfloat[Experimental setup.]{
\label{fig:improved_subfig_b}
\begin{minipage}[t]{0.5\textwidth}
\centering
\includegraphics[width=80mm, clip]{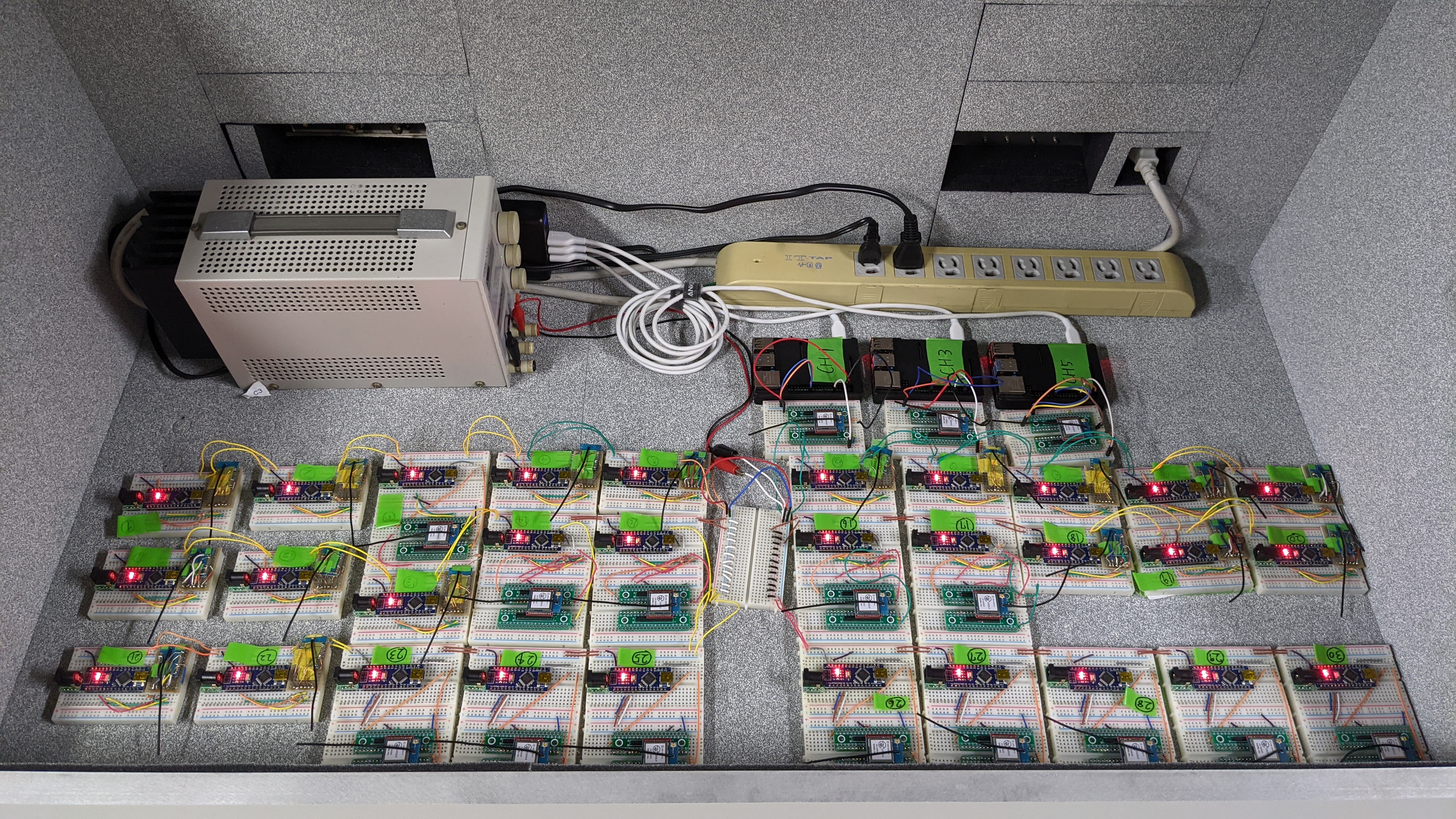}
\end{minipage}
}
\caption{Experimental settings.}
\end{figure}
In our implementation, we consider a practical LoRaWAN system that comprises one multichannel GW and multiple LoRa devices, as shown in Fig. 4. 
Fig. 4 (a) shows the constitution of the GW and LoRa devices in the LoRaWAN system built in this study.
As shown in Fig. 4 (a), multichannel GW is fabricated using Raspberry Pi and ES920LR. ES920LR is a LoRa-modulation based wireless module working on 920 MHz band, while Raspberry Pi is a controller that regulates the LoRa module ES920LR and save data and send to the sever for analysis. Each set of Raspberry Pi and ES920LR is used to realize one available channel of the GW. 
The parameters used by ES920LR can be optimized according to the user application since its LoRa specifications is original private.
Each LoRa device comprises one Arduino nano and one ES920LR, where Arduino nano is a controller to running the lightweight transmission parameter selection schemes. 
The experiments are conducted in a shield box to avoid interference from other devices working on 920 MHz. The shield box used in our experiments is MY1530 developed by MICRONIX Co.,LTD. The experimental setup is shown in Fig. 4 (b).


\subsection{Compared Schemes}
In this subsection, we present the compared schemes, i.e., upper confidence bound (UCB1)-Tuned based, $\epsilon$-greedy-based, and random transmission parameters selection schemes. 
The reason for comparing these three schemes to our proposed scheme is that all of the three schemes are lightweight decentralized learning methods that can be implemented on IoT devices with limited computational ability and memory. In addition, UCB1-tuned and $\epsilon$-greedy approaches are state-of-the-art MAB algorithms. The effectiveness of our proposed scheme can be displayed well through the comparison with these two methods.

\subsubsection{UCB1-tuned-based transmission parameter selection}
UCB1-tuned algorithm belongs to the UCB family, where the arm is selected by calculating a "UCB" given a history of rewards and a number of pulls for each arm. The simplest algorithm in the UCB family is the UCB1 algorithm where the arm is selected based on empirical means. 
It is shown that optimal regret up to a multiplicative constant can be achieved using the UCB1 algorithm \cite{18}. That is, UCB1 is efficient to solve MAB problems.
The UCB1-tuned algorithm is an enhanced version of the UCB algorithm. In the UCB1-tuned algorithm, arm variance is also considered along with the empirical means. It is presented that the UCB1-tuned algorithm can obtain better performance than the UCB algorithm in practice.
In the UCB1-tuned-based transmission parameter selection scheme, the LoRa device selects its access channel and SF at the $t$-th decision by
\begin{equation}
    \tiny{ k_j^{\ast}(t) =\mathrm{argmax}_{k_j\in CH_U or SF} \left(\hat{\mu_{k_j}}+\sqrt{\frac{\ln t}{N_{k_j}(t)}\min\left(\frac{1}{4},V_{k_j}(N_{k_j}(t))\right)}\right)},
\end{equation}
where $\hat{\mu_{k_j}}$ is the empirical mean of the obtained reward for arm $k_j$. In addition, the expression of the $V_{k_j}(N_{k_j}(t))$ is given as follows:
\begin{equation}
    V_{k_j}(N_{k_j}(t)) = \hat{\sigma_{k_j}^2} + \sqrt{ \frac{2 \ln t}{N_{k_j}(t)}},
\end{equation}
where $N_{k_j}(t)$ is the number of times that machine $k_j$ has been pulled until time $t$ and $\hat{\sigma_{k_j}^2}$ is the variance of the arm $k$, which can be calculated as usual.

\subsubsection{$\epsilon$-greedy-based transmission parameter selection} In the $\epsilon$-greedy-based parameter selection scheme, the transmission parameters, i.e., channel and SF, are selected randomly with probability $\epsilon$, while the channel or SF with the highest probability of successful transmissions is selected with probability $1-\epsilon$. The expression of this scheme is given as follows: 
\begin{equation}
     k_j^{\ast}(t) =
    \begin{cases}
    \mathrm{argmax}_{k_j \in {CH_U} \text{or} SF} {X_{k_j}^g}(t), \text{with probability} \ (1-\epsilon),\\
    \text{randomly select CH and SF},
    \ \text{with probability $\epsilon$},
    \end{cases}
    \label{eq:n_ik}
\end{equation}
where ${X_{k_j}^g}(t)$ denotes the probability of successful transmission using channel $k_j$ at the $t$-th decision-making.  
\subsubsection{random transmission parameter selection} In the random transmission parameter selection scheme, the channel and SF are selected randomly by the LoRa device.

The learning parameters used in our experiments are set as constants. The learning parameters $\alpha$, $\beta$, and $A$ in the proposed scheme is set to 0.9, 0.9, and 0.5, respectively. 
The $\epsilon$ in the $\epsilon$-greedy-based transmission parameter selection scheme is set to 0.1.


\subsection{Performance Evaluation of the Channel Selection}

In this subsection, we evaluate the FSR of the built LoRaWAN system for the channel selection using the proposed scheme with/without varying available channels. First, we present the common parameter settings used in the experiments presented in this section. 
Then, we evaluate the FSR of the built LoRaWAN system for the channel selection using our proposed approaches without/with varying available channels, and then we compare it to the UCB1-tuned-based, $\epsilon$-greedy-based, and random transmission parameter selection schemes. 

\subsubsection{Parameter Settings}
The bandwidth and frequency used in the experiments are 125 KHz and 920 MHz, respectively. The number of available channels for GW and LoRa devices are 3 and 5, respectively. In the experiments presented in this section, each LoRa device selects its access channel from the five available channels using lightweight transmission parameter selection schemes. The SF used by LoRa devices is set to 7. The transmission interval $TI$, payload $L$, and re-transmission time $TR$ are set to 10 s, 50 bytes, and 3, respectively. The transmission power used by the LoRa devices is set to 20 mW.
The values of the parameters 
are summarized in Table. II.

\begin{table}[htbp]
	\centering 
	\caption{Parameter settings for channel selection.}  
	\label{table1} 
	\begin{tabular}{|c|c|} 
	\hline 
	Parameter&Value\\ 
	\hline
	Frequency & 920MHz\\ 
	\hline
	Bandwidth $B$&125kHz\\ 
	\hline
		Channels used by each GW&CH1,CH3,CH5\\ 
	\hline
	Channels used by LoRa devices &CH1,CH3,CH5, CH7, CH9\\ 
	\hline
	SF&7\\ 
	\hline
Transmission interval $TI$&10 s\\ 
	\hline
    Payload $L$&50 B\\ 
	\hline
	Retransmission time $TR$&3\\ 
	\hline
	Transmission power of the LoRa devices&20 mW\\ 
	\hline
	\end{tabular}
\end{table}

\subsubsection{Scenario 1: LoRaWAN system without varying available channels}

\begin{figure}
    \centering
    \includegraphics[width=75mm, height=70mm]{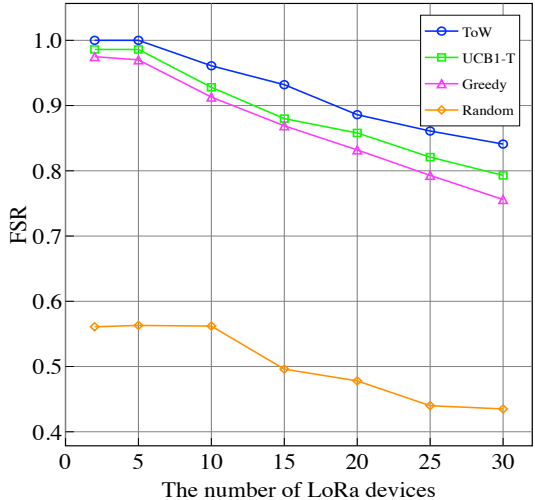}
    \caption{FSR for the channel selection in scenario 1.}
    \label{fig:systemmodel}
\end{figure}
In this subsection, we evaluate the FSR performance for channel selection 
using the lightweight transmission parameter selection schemes
in LoRaWAN without varying available channels. 
In the experiments, the number of LoRa devices are set to 2, 5, 10, 15, 20, 25, and 30, respectively.
The experimental results are shown in Fig. 5.
The FSR shown on the y-axis in Fig. 5 is the average FSR during the 30 min experiments.  
As shown in Fig. 5, as the number of LoRa devices in LoRaWAN increases, the FSR decreases for any of the evaluated schemes.
A possible reason may be that with the increase in the number of LoRa devices in LoRaWAN, more LoRa devices unavoidably access the same channel simultaneously. Hence, the collision probability between LoRa devices increases.
In addition, the FSR when access channels are selected using MAB-based schemes, i.e., UCB1-tuned-based and $\epsilon$-greedy-based transmission parameter selection schemes, is much higher than random selection schemes irrespective of the number of LoRa devices used in LoRaWAN. The reason may that MAB-based schemes can learn the environment and dynamically adjust the selected channels.
Moreover, selecting channels using our proposed ToW dynamics-based scheme can achieve higher FSR than the other MAB-based schemes, especially when the number of LoRa devices become larger. Since LoRa devices have a long distance communication ability, an increased amount of LoRa signals is concentrated in the same area. Hence, the proposed scheme is more effective for LoRaWAN than other MAB-based schemes.  

\begin{figure*}[htbp]
\centering
\subfloat[ToW Dynamics]{
\label{fig:improved_subfig_a}
\begin{minipage}[t]{0.24\textwidth}
\centering
\includegraphics[width=1.6in]{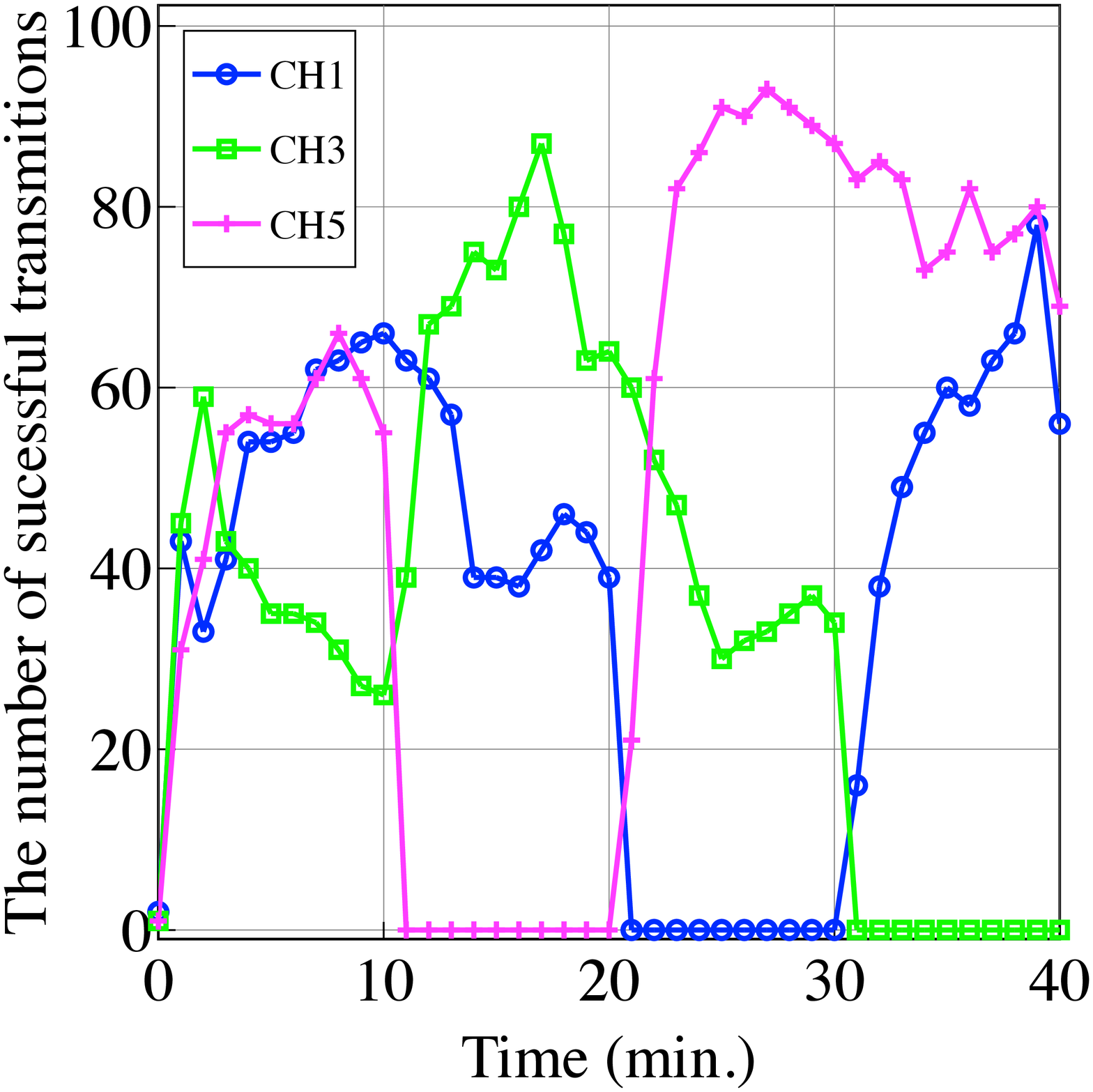}
\end{minipage}
}
\subfloat[UCB1-tuned]{
\label{fig:improved_subfig_b}
\begin{minipage}[t]{0.24\textwidth}
\centering
\includegraphics[width=1.65in]{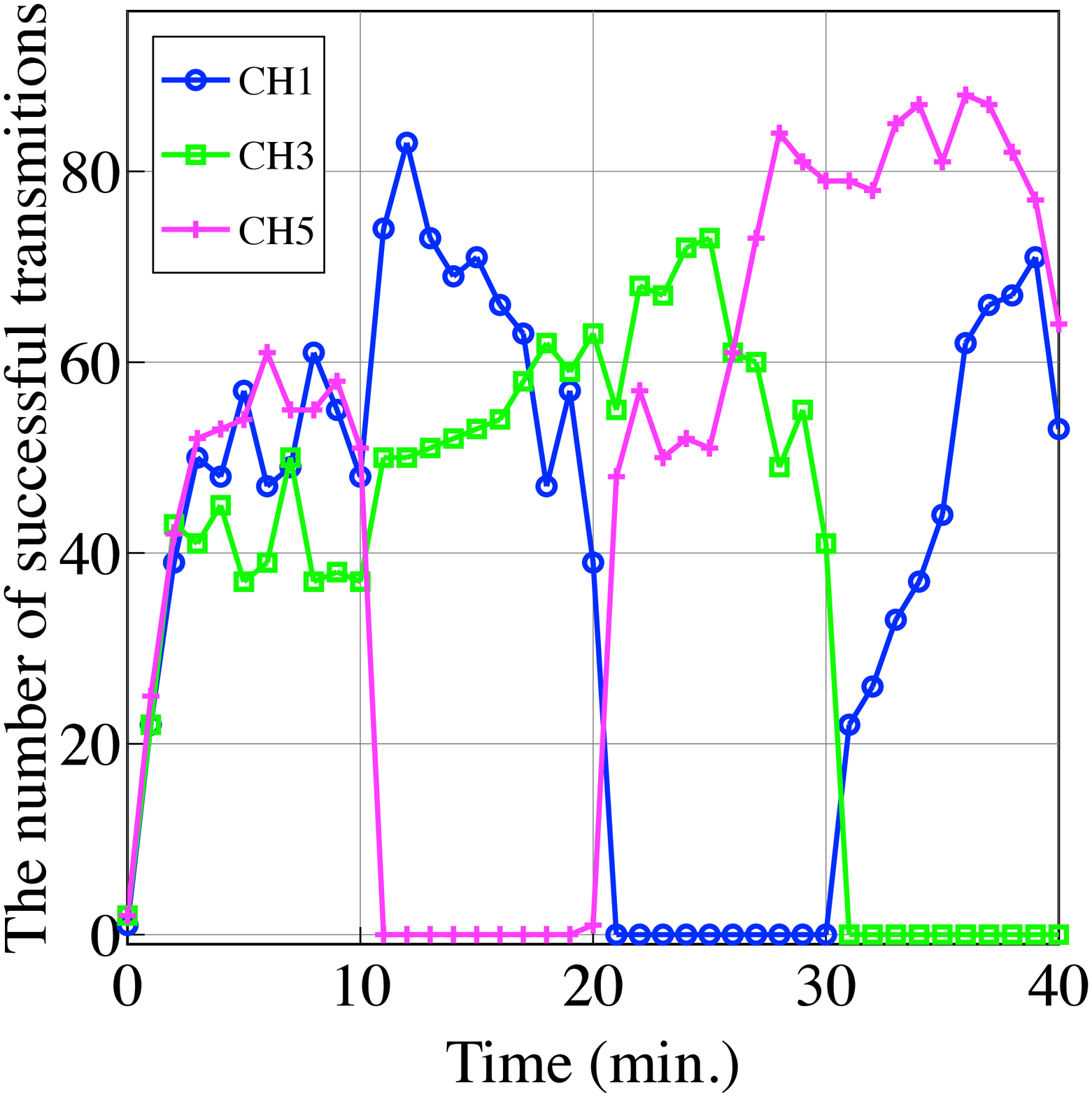}
\end{minipage}
}
\subfloat[$\epsilon$-greedy]{
\label{fig:improved_subfig_b}
\begin{minipage}[t]{0.24\textwidth}
\centering
\includegraphics[width=1.65in]{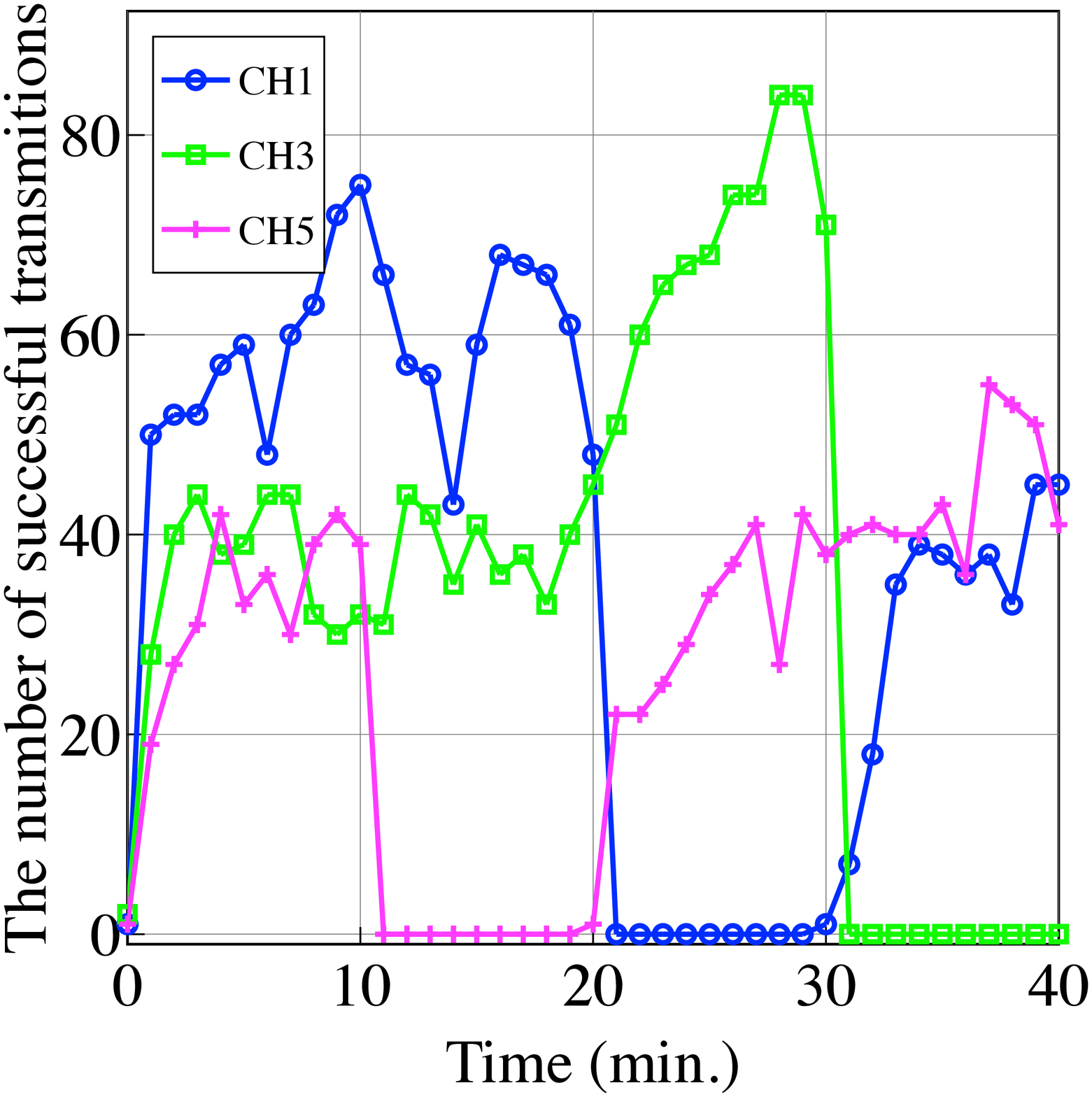}
\end{minipage}
}
\subfloat[Random]{
\label{fig:improved_subfig_b}
\begin{minipage}[t]{0.24\textwidth}
\centering
\includegraphics[width=1.65in]{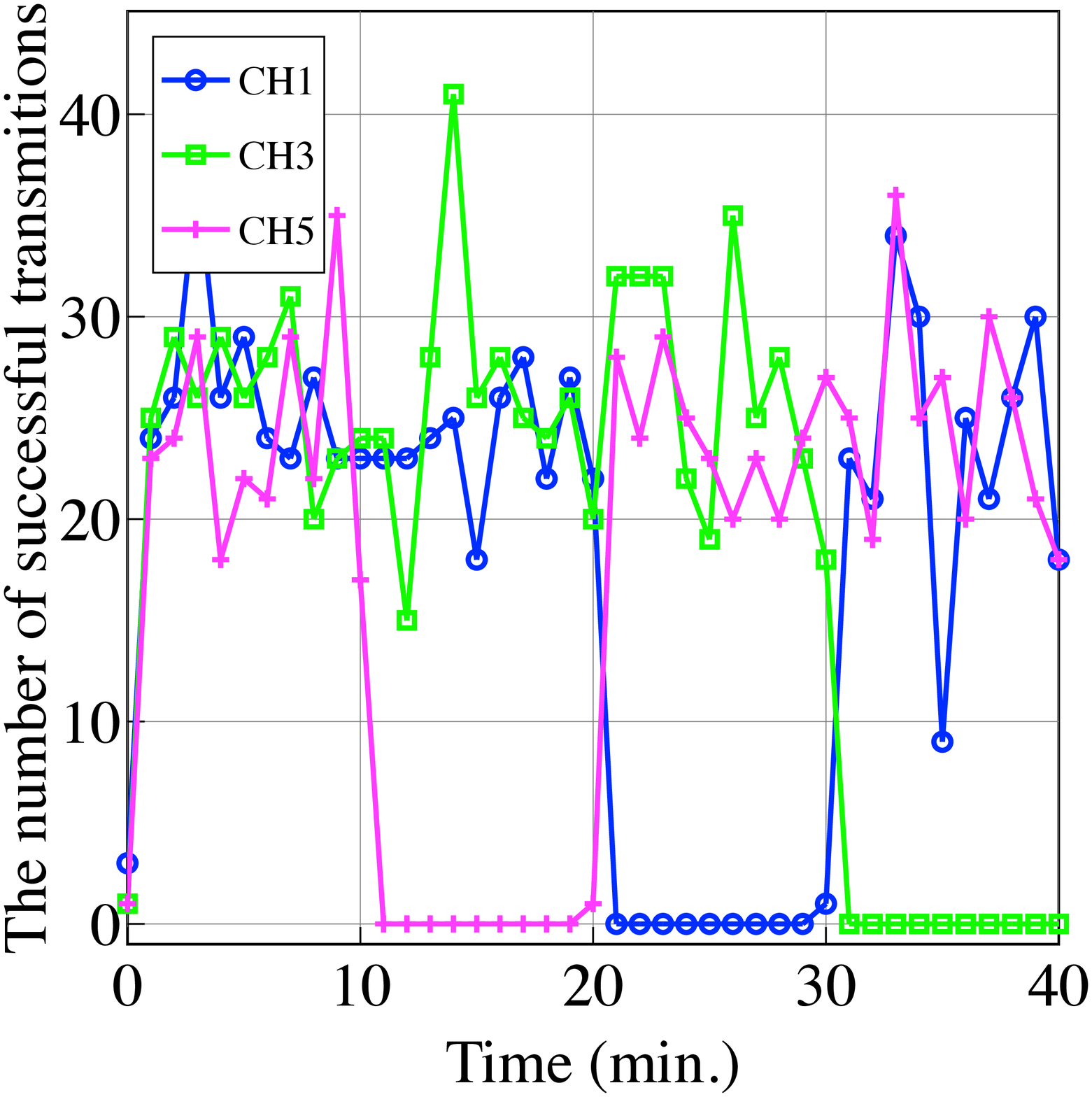}
\end{minipage}
}
\caption{Number of successful transmissions during the experiments.
}
\end{figure*}

 \begin{figure}
    \centering
    \includegraphics[width=70mm, height=45mm]{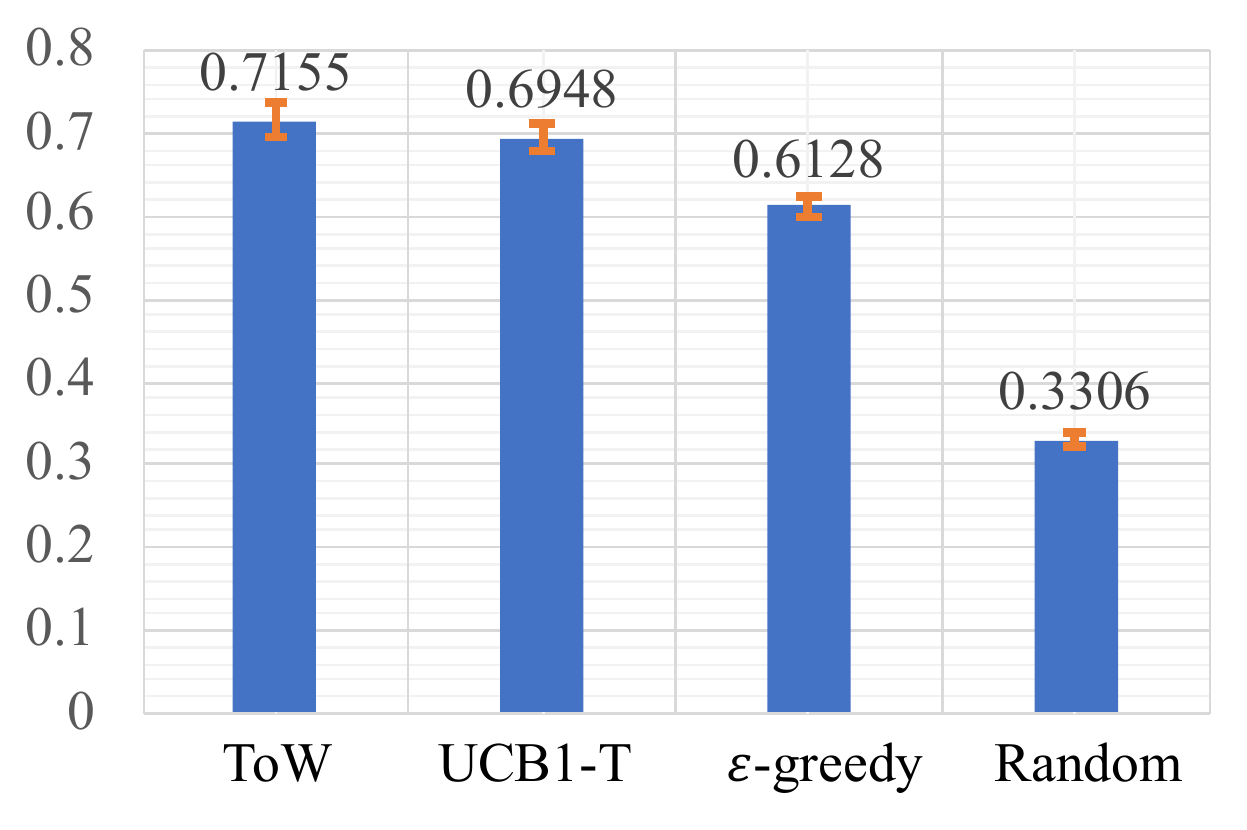}
    \caption{Average FSR for the channel selection in scenario 2.}
    \label{fig:systemmodel}
\end{figure}

\subsubsection{Scenario 2: LoRaWAN system with varying available channels} In this subsection, the FSR performance under the LoRaWAN with varying available channels is evaluated. In this experiment, the number of LoRa devices is set to 30. The available channels of GW are changing dynamically during the experiment, as shown in Table III .
That is, all of the three available channels for GW are available during 0-10 min. CH5, CH1, and CH3 are not available during 10-20 min, 20-30 min, 30-40 min, respectively. 
To make the channel unavailable, Raspberry Pi is used to control the LoRa module not to receive data from corresponding channels.

\begin{table}[htbp]
	\centering 
	\caption{Settings of the available channels.}
	\label{table1} 
	\begin{tabular}{|c|c|}
	\hline 
	Time (min.) &Available channels\\ 
	\hline
	0-10 & CH1, CH3, CH5\\ 
	\hline
	10-20&CH1, CH3\\ 
	\hline
    20-30&CH3, CH5\\ 
	\hline
	30-40&CH1, CH5\\
	\hline
	\end{tabular}
\end{table}

The number of successful transmissions during the experiments within 40 min is shown in Fig. 6, indicates that the proposed ToW dynamics-based scheme can adjust the access channel quickly when the available channel changes. 
That is, the channels that were not available in the previous duration but are available in the current duration can be selected more quickly using the proposed scheme as compared to other MAB algorithm-based schemes.
The reason may be that the proposed scheme can learn the environment and adjust its decisions quickly.
Hence, the proposed scheme may be more suitable for dynamic environments.
Next, we conduct the above experiments 10 times to obtain the average FSR for the channel selection in this scenario 2. 
The experimental results are shown in Fig. 7.
We can observe in the figure that our proposed scheme can achieve higher average FSR than the other schemes.
In addition, the confidence interval for the ToW, UCB1-tuned, $\epsilon$-greedy, and random-based transmission parameter selection schemes are 0.021, 0.018, 0.012, and 0.007, respectively. The confidence level is set to 95\%. 
The corresponding standard deviations are 0.030, 0.025, 0.016, and 0.010, respectively. Although the confidence interval and standard deviation of the proposed scheme are larger than the other schemes, the difference is small.
Moreover, we observe that the FSR in this scenario is much lower than that in LoRaWAN without varying the available channels. The reason may be that 
we stopped the data transmission through one channel from 10 to 40 min to set the dynamic environment.
That is, the number of available channels is changed from 3 to 2 during 10-40 min compared to the LoRaWAN system without varying the available channels, thereby exacerbating the channel collisions among the LoRa devices.
Furthermore, the FRS is not only influenced by the channel but also by several other factors, such as SF, transmission interval, the interval of neighboring channels, and transmission power. In the remainder of this section, we evaluate the interference on the FSR by SF and the interval of neighboring channels.
\subsection{Performance Evaluation of the CH-SF Selection}

In this subsection, we evaluate the performance of the CH-SF selection using the proposed scheme, and compare it to UCB1-tuned-based and $\epsilon$-greedy-based CH-SF selection approaches. First, we present the parameter settings used in the experiment. Then, we discuss the performance of the CH-SF selection without available channel changes. Finally, to evaluate our proposed approach in a more realistic scenario, we evaluate our proposed approach under the scenario where the LoRa devices coexist with other types of IoT devices. In our experiment, we use the wireless smart utility network (Wi-SUN) devices as the other type of IoT devices. Similar to LoRa devices, Wi-SUN is used in the 920 MHz band in Japan. It also has the advantages of long communication distance, low power consumption, etc \cite{40}.  
\subsubsection{Parameter Settings}
In the experiment, the settings of the frequency, bandwidth, payload, and transmission power of the LoRa devices are same as those in the last subsection, i.e., 920 MHz, 125 KHz, 50 B, and 20 mW, respectively.
The numbers of available channels for the GW and devices are set to 3, i.e., CH1, CH3, and CH5. 

SFs that can be used by LoRa devices are set to 7, 8, and 9. Since each CH-SF pair of the GW is relized by one controller Raspberry Pi and one LoRa module ES920LR, the number of the Raspberry Pi-ES920LR device sets is 9, which equals to the number of combinations of the CH-SF pair (i.e., $3\times3$). The number of LoRa devices is set to 30.
Each LoRa device accesses the channel using the SF selected by the proposed scheme.
Since the on-air time for SF 8 and SF 9 are much longer than that of SF 7, to satisfy the duty ratio of the LoRa devices, the transmission interval is set to 20 s.
Moreover, to reduce the performance influences by retransmission, the retransmission time is set to 0 here. 
The total number of transmission attempts of LoRa devices in this experiment is set to 200.
Table IV summarizes the main parameter settings. 

\begin{table}[htbp]
	\centering 
	\caption{Parameter settings for the CH-SF selection.} 
	\label{table1} 
	\begin{tabular}{|c|c|} 
	\hline 
	Parameter&Value\\ 
	\hline
    Number of devices $M$&30\\ 
		\hline
		Channels used by GW&CH1,CH3,CH5\\ 
	\hline
	Channels used by LoRa devices &CH1,CH3,CH5\\ 
	\hline
	SF&7, 8, 9\\ 
	\hline
Transmission interval $TI$&20 s\\ 
	\hline
	Retransmission time $TR$&0\\ 
		\hline
	Times of transmission attempts&200\\ 
	\hline
	\end{tabular}
\end{table}

 \begin{figure}
    \centering
    \includegraphics[width=70mm, height=45mm]{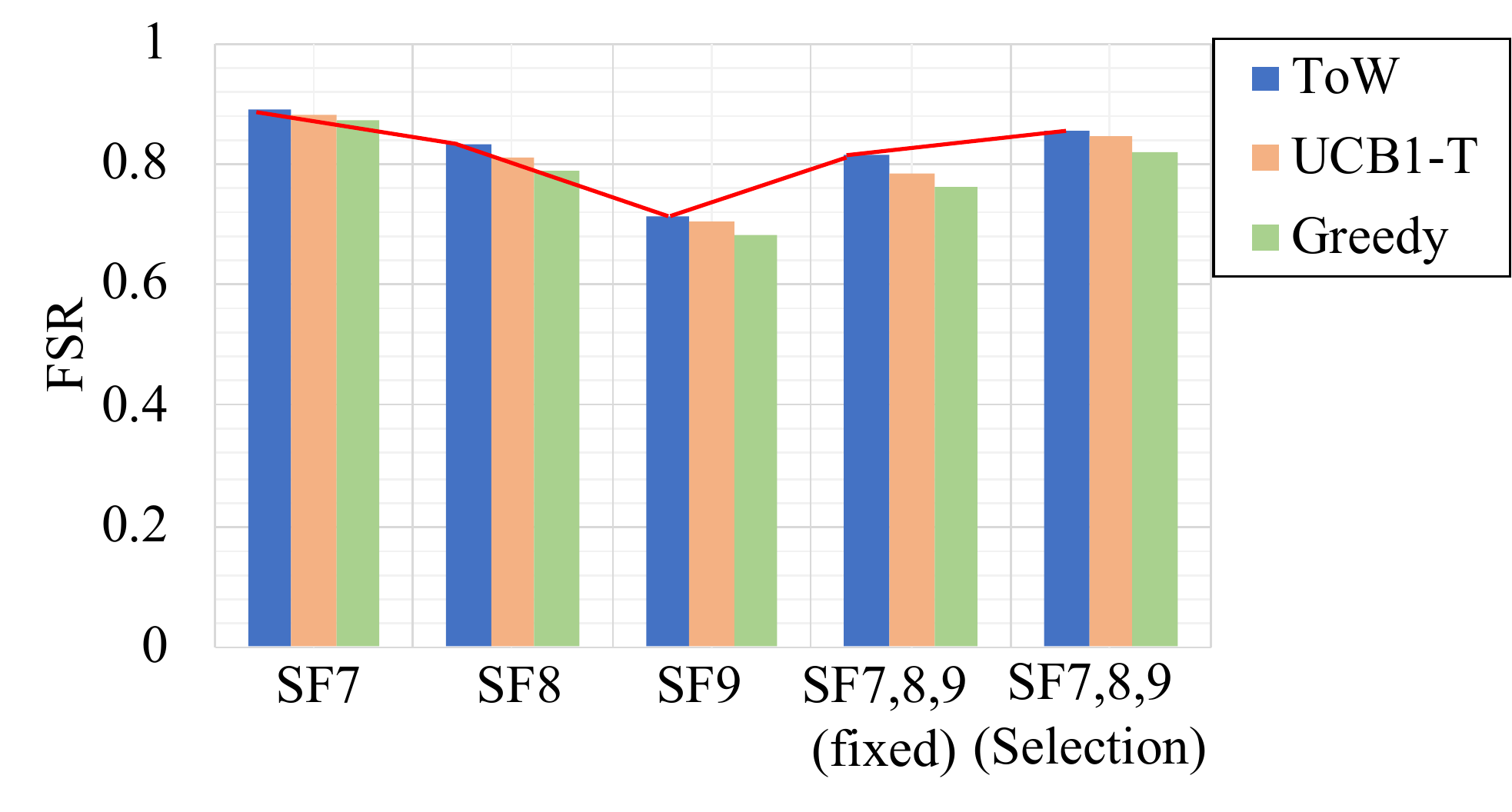}
    \caption{FSR for CH-SF selection in scenario 1.}
    \label{fig:systemmodel}
\end{figure}

\subsubsection{Scenario 1: High-density LoRaWAN system without varying available channels}
We evaluate the FSR performance for CH-SF selection using our proposed scheme and compare it to the UCB1-tuned-based and $\epsilon$-greedy-based CH-SF selection approaches. 
We evaluate five settings in the experiments.
Each device can select an access channel from CH1, CH3, and CH5 for all of the settings. The difference among the five settings is the SF. 

For settings 1, 2, and 3, the SF used by LoRa devices are fixed to 7, 8, and 9, respectively. 
For setting 4, the LoRa devices are equally divided into three groupsm, and each group consists of 10 LoRa devices. The SF for each group is fixed to 7, 8, and 9, respectively.
For setting 5, each device selects SF from 7, 8, or 9 based on lightweight learning schemes. The settings described above are summarized in Table V. 

\begin{table}[htbp]
	\centering 
	\caption{Settings of the CH-SF selection.}
	\label{table1} 
	\begin{tabular}{|c|c|}
	\hline 
	Number &Settings\\ 
	\hline
	1& $SF=7$, channel selection\\ 
	\hline
	2&$SF=8$, channel selection\\ 
	\hline
    3&$SF=9$, channel selection\\ 
	\hline
	4&$SF=7, 8, 9$, channel selection\\
	\hline
	5&CH-SF selection\\
	\hline
	\end{tabular}
\end{table}

The experimental results are shown in Fig. 8. The experimental results is the average FSR of 10 trials of the experiment. 
Fig. 8 shows that the FSR decreases with the increase in the value of SF. This is because the time on air increases as the value of SF increases, which increases the collisions among LoRa devices. Moreover, the receiver sensitivity is high enough for successful communications when $SF=7$, since the distance between LoRa devices and SF is very small in our setting. Hence, we can say that $SF=7$ is the optimal SF in our experimental settings. 
Additionally, we can observe that the FSR for CH-SF selection is near the FSR when the SFs for LoRa devices are fixed to 7, showing that LoRa devices can select appropriate SF based on our proposed scheme. Moreover, from the experimental results of settings 4 and 5, i.e., channel selection and CH-SF selection, we can observe that CH-SF selection 
can improve the FSR compared to channel selection only. Hence, FSR can be improved by selecting optimal CH-SF. Furthermore, our proposed ToW dynamics-based transmission parameters selection scheme can achieve the highest FSR among the lightweight schemes.

\subsubsection{Scenario 2: High-density LoRaWAN system with varying available channels}

\begin{figure}
    \centering
    \includegraphics[width=70mm, height=40mm]{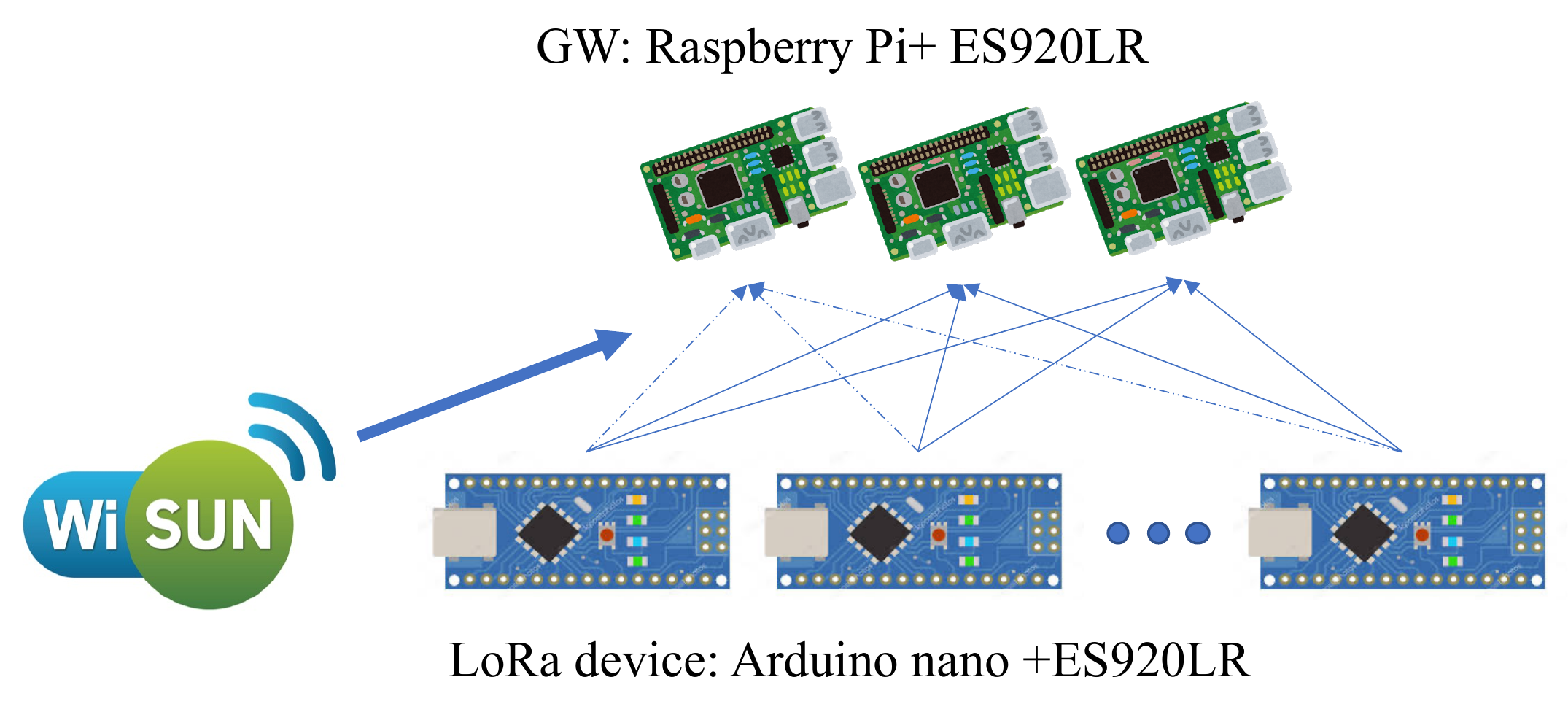}
    \caption{Experimental setting for the CH-SF selection in scenario 2.}
    \label{fig:systemmodel}
\end{figure}

In this subsection, we evaluate the performance in a realistic communication environment where the LoRaWAN system coexists with other IoT systems. In our experiment, we set the Wi-SUN system as the other IoT system. The experimental setting is shown in Fig. 9. The parameter settings of Wi-SUN is summarized in Table VI. As shown in the table, the bit rate, transmission interval, payload, the number of devices, retransmission time, and transmission power of Wi-SUN devices are set to 50 kbps, 1 s, 200 bytes, 20, 0, and 20 mW, respectively. In addition, to imitate a realistic scenario while evaluating the performance of our proposed scheme in dynamic environments,
the access channels of Wi-SUN devices are varied, as shown in Table VII. That is, Wi-SUN devices do not transmit data during 0-20 min while transmitting data using channel 1 during 20-30 min. Then, Wi-SUN devices change the access channel to channel 3 during 30-40 min. The access channel of the Wi-SUN devices change to channel 5 during 40-50 min. Wi-SUN devices change to use channel 1 during 50-60 min. As describe above, the devices change their access channels sequentially every 10 minutes.

\begin{table}[htbp]
	\centering 
	\caption{Parameter settings of Wi-SUN devices.}
	\label{table1} 
	\begin{tabular}{|c|c|}
	\hline 
	bit rate & 50 kbps\\ 
	\hline
	Transmission interval&1 s\\ 
	\hline
    Payload&200bytes\\
	\hline
	Number of devices&20\\
	\hline
	Retransmission time&0\\
	\hline
	Transmission power&20 mW\\
	\hline
	Access channels& CH1->CH3->CH5\\ 
	\hline
	\end{tabular}
\end{table}

\begin{table}[htbp]
	\centering 
	\caption{Access channel of Wi-SUN devices.}
	\label{table1} 
	\begin{tabular}{|c|c|}
	\hline 
	Time (min.) &Available channels\\ 
	\hline
	0-20 & ---\\ 
	\hline
	20-30&CH1\\ 
	\hline
    30-40&CH3\\ 
	\hline
	40-50& CH5\\
	\hline
	\end{tabular}
\end{table}

\begin{figure}
    \centering
    \includegraphics[width=70mm, height=45mm]{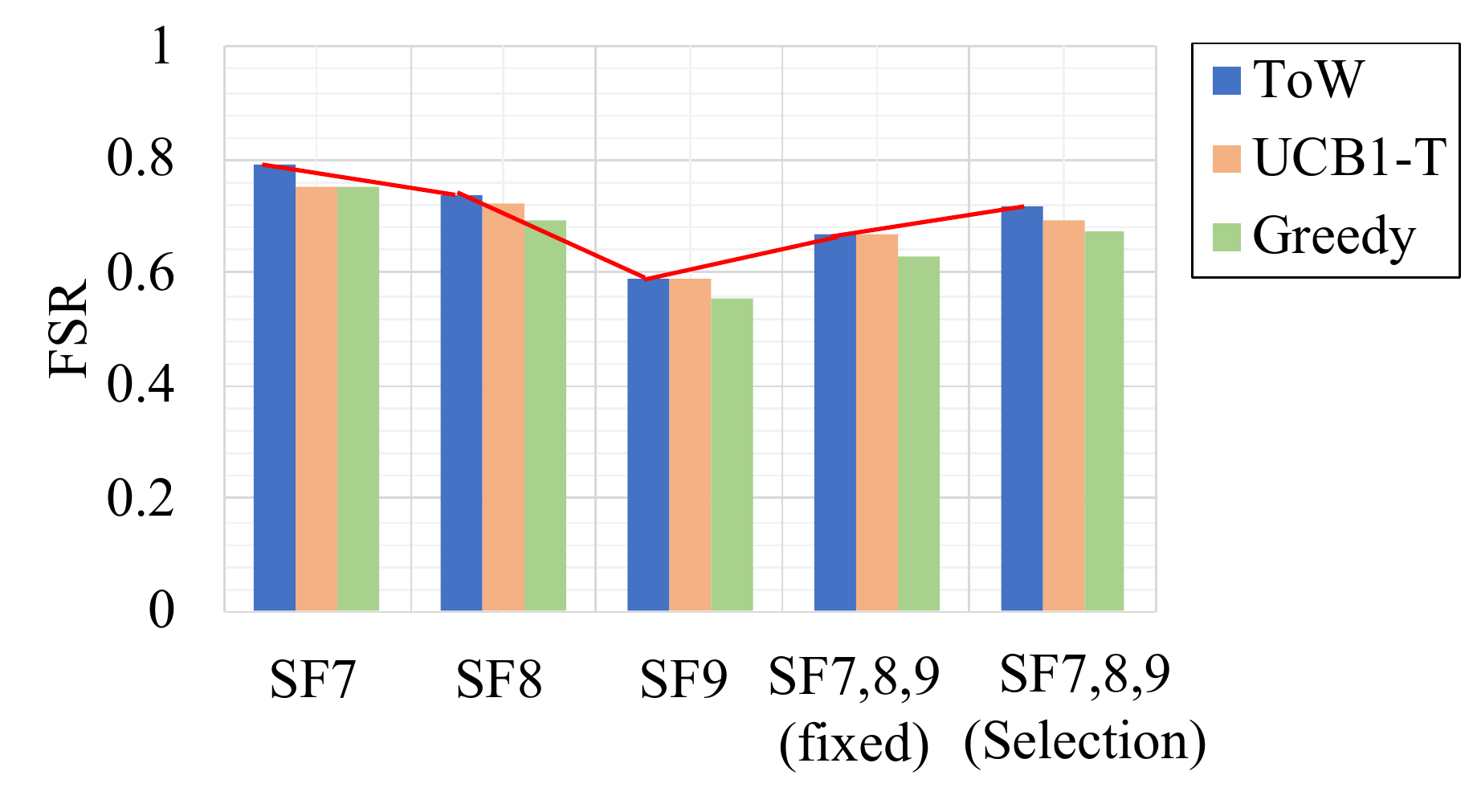}
    \caption{FSR for the CH-SF selection in scenario 2.}
    \label{fig:systemmodel}
\end{figure}

Fig. 10 shows the experimental results of selecting CH-SF under the above settings.
Fig. 10 depicts the average FSR of 10 trials of the experiment. The parameter settings of the LoRa devices are presented in Table IV.
From Fig. 10, we can observe that FSR decreases as SF increases when there is no variation in available channels. This is because the on-air time increases when SF increases, which may increase the collisions among LoRa devices. Moreover, we can observe that our proposed ToW dynamics-based CH-SF selection scheme can achieve higher FSR than the other schemes, while the CH-SF selection using our proposed scheme trends to the optimal setting, i.e., SF=7. Furthermore, similar to the experimental results of the system with variations in the available channels, the FSR can be improved in the system with varying available channels by selecting the CH-SF pair compared to only selecting the channel.

\subsection{Interference Between Adjacent Channels}
In this subsection, we evaluate the performance on interference between adjacent channels.
To evaluate this, the available channels are set to $\{CH2, CH4, CH6\}$ and $\{CH2, CH5, CH8\}$, respectively. The number of LoRa devices is set to 30. In addition to the channel and number of LoRa devices, other parameters are presented in Table II. 
We evaluate the FSR and fairness performance to verify the interference between adjacent channels. The FSR can be calculated using equation (3), while fairness \cite{41} can be expressed as

\begin{equation}
    FI=\frac{(\sum_{k_1=1}^{I}R_{k_1})^2}{I\sum_{k_1=1}^{I}R_{k_1}^2}.
\end{equation}

The experimental results of fairness and the FSR for the random-based channel selection and ToW dynamics-based channel selection are presented in Table VIII.
In the table, the settings Random\_1 and ToW\_1 correspond to the channel set $\{CH2, CH4, CH6\}$, while Random\_2 and ToW\_2 correspond to the channel set $\{CH2, CH5, CH8\}$. 1CH, 2CH, and 3CH denote the first, second, third channels in the available channel set. 
From Table VIII, we can observe that the packets received by the second channel is significantly less than the other channel in Random\_1 and ToW\_1. This is because interference occurs between two adjacent channels. To improve fairness and the FSR of each channel, we set the available channels with interval of two channels, i.e., CH2, CH5, and CH8. The results show that fairness increases from 92.6\% to 98.8\% and from 99.8\% to 99.9\% for the ToW dynamics-based channel selection scheme and the random-based channel selection scheme, respectively. Meanwhile, the FSR increases from 80.7\% to 86.8\% and from 73.3\% to 75.9\% for the ToW dynamics-based channel selection scheme and random-based channel selection scheme, respectively.  
Hence, we can assume that the FSR and fairness can be improved by increasing the interval of adjacent channels.

\begin{table}[]
    \caption{Fairness and FSR for different CH intervals.}
    \vspace{20pt}
    \centering
    \begin{tabular}{p{1cm}p{1cm}p{1cm}p{1cm}p{1cm}p{1cm}p{1cm}}
        \hline
     \multirow{2}*{Settings} &\multicolumn{3}{c}{Number of received packets (\%)}&\multirow{2}*{FSR}&\multirow{2}*{Fairness}\\
     \cline{2-4}&1CH&2CH&3CH 
     \\
        \hline
         Random\_1 & 1496 (33.8\%) & 1398 (31.6\%) & 1398 (34.7\%) & 73.3\% & 99.8\%\\
           \hline
          ToW\_1 & 1902 (38.5\%) & 998 (20.2\%) & 2045 (41.4\%) & 80.7\% & 92.6\%\\
        \hline
         Random\_2 & 1591 (34.5\%) & 1545 (33.5\%) & 1470 (31.9\%) & \textcolor{blue}{75.9\%} & \textcolor{blue}{99.9\%}\\
        \hline
         ToW\_2 & 1494 (28.2\%) & 1892 (35.8\%) & 1906 (36\%) & \textcolor{blue}{86.8\%} & \textcolor{blue}{98.8\%}\\
        \hline
    \end{tabular}
    \label{bs2}
\end{table}

Then, we evaluate the FSR of the channel selection in a dynamical available channel scenario, where LoRa devices coexists with Wi-SUN devices. The available channels for GW are set to CH1, CH4, and CH7, while that for LoRa devices are set to CH1, CH4, CH7, CH10, and CH14. The settings, except for the available channels, are the same as in TABLE II. The access channels of Wi-SUN devices are set as follows. Wi-SUN devices do not access any channel during 0-10 min, but they access CH1, CH4, and CH7 during 10-20 min, 20-30 min, 30-40 min, respectively. The other parameter settings of Wi-SUN are the same as in TABLE VI.
Fig. 11 shows the experimental results of the average FSR over 10 trials of the experiment.
Figure 11 shows that the FSR decreases with an increase in the number of LoRa devices. This is because the probability of collisions among the LoRa devices increases with an increase in the number of LoRa devices. Moreover, our proposed ToW dynamics-based channel selection scheme can achieve the highest FSR among the three schemes.
The comparison of the average FSR when the channel interval is 1 and 2 is shown in Fig. 12. The figure shows that the FSR can be improved to a great extent when the channel interval is 2 as opposed to when it is 1.

\begin{figure}
    \centering
    \includegraphics[width=75mm, height=70mm]{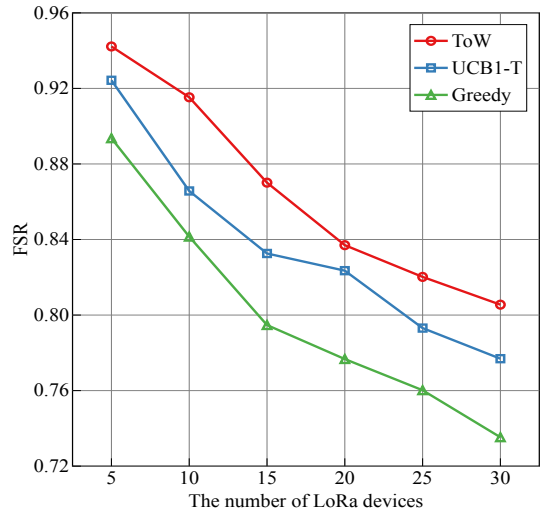}
    \caption{FSR when the CH interval is 2.}
    \label{fig:systemmodel}
\end{figure}

\begin{figure}
    \centering
    \includegraphics[width=65mm, height=30mm]{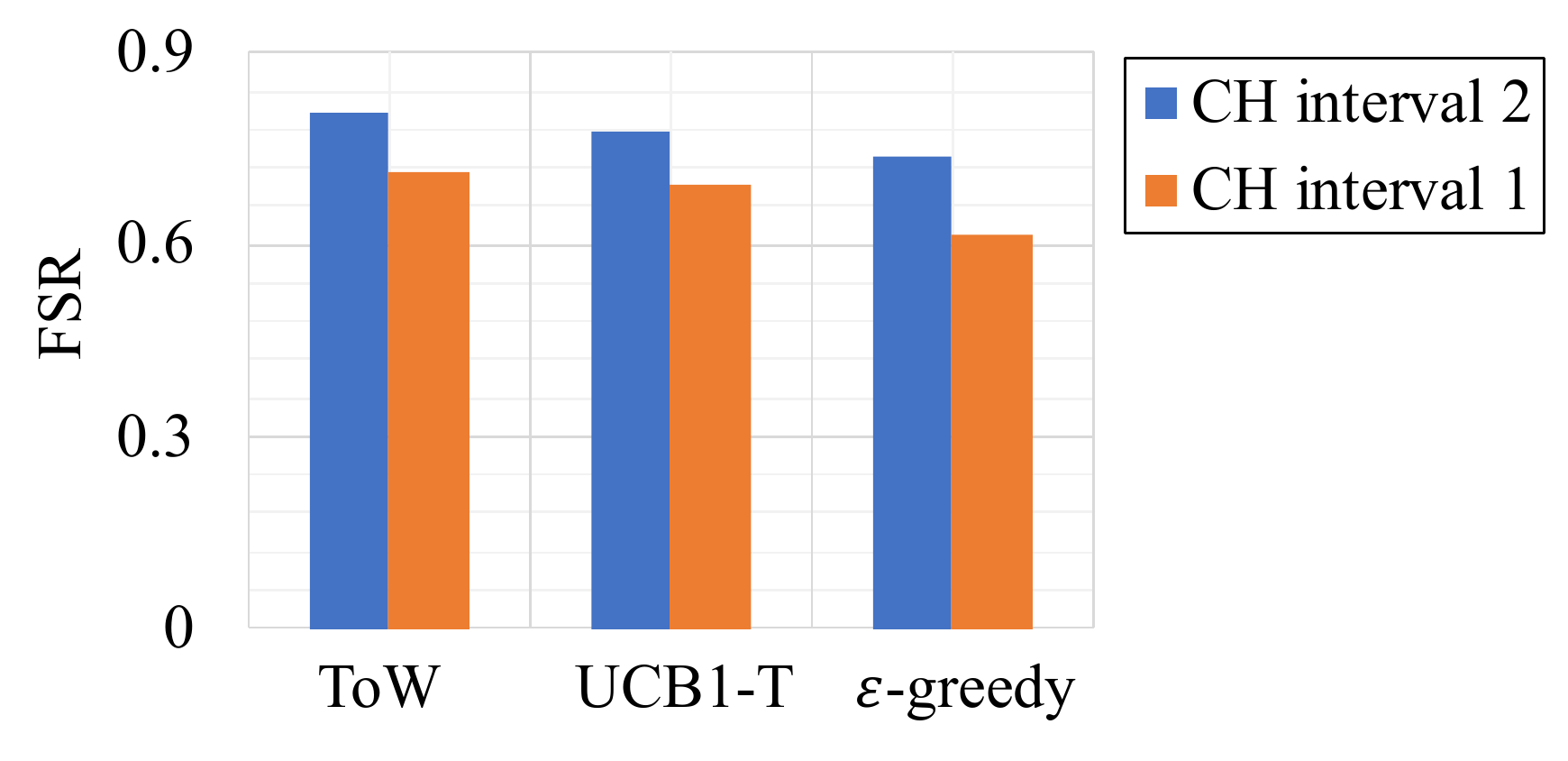}
    \caption{Comparison of the FSR for different CH intervals.}
    \label{fig:systemmodel}
\end{figure}



\section{Conclusion}
\label{sect:experiment}
This paper proposed a lightweight transmission parameters selection scheme using reinforcement learning to avoid collisions between LoRa devices. 
The proposed scheme can select appropriate channels and SFs based only on the ACK information using simple four arithmetic operations. 
Theoretical analysis showed the extremely low computational complexity and memory requirement of the proposed scheme. 
Hence, it is easy to implement it on practical LoRa devices with limited computational ability and memory. 
To evaluate the effectiveness of our proposed scheme, we implemented the proposed scheme on practical LoRa devices and conducted several experiments on the built LoRaWAN. Experimental results showed the following phenomena: 
(1) Compared to other lightweight transmission parameter selection schemes, collisions between LoRa devices can be more efficiently avoided by our proposed scheme in LoRaWAN irrespective of the changes in the available channels.
(2) FSR can be improved by jointly selecting channels and SFs as opposed to only selecting channels. (3) Since interference exists between adjacent channels, the FSR and fairness can be improved by increasing the interval of adjacent available channels.

In addition to channel and SF selection, some other transmission parameters of LoRa devices will influence the communication performance (e.g., transmission power and transmission interval). Larger transmission power may increase the strength of the signal. Meanwhile, the energy efficiency may also be increased. A larger transmission interval may reduce the collisions among IoT devices, while communication latency may be increased. Hence, it is crucial to select appropriate transmission power, transmission interval, and other transmission parameters, which will be considered in our future work.
Moreover, energy efficiency will be considered. Although MAB-based approaches can achieve higher FSR than the random approach, the energy consumption of the random approach may be much lower than MAB-based approaches. There should be a tradeoff between FSR and energy efficiency. For IoT devices powered by batteries and arranged in hard-to-reach places, energy efficiency may be more important than FSR. Meanwhile, the FSR is extremely important for low latency tolerant applications, such as vehicular communications. Hence, in our opinion, selecting different methods for different
applications is an extremely interesting topic and can be explored as future scope of this study.


\begin{thebibliography}{00}

\bibitem{5}
Y. Liu, J. Wang, J. Li, S. Niu and H. Song, "Machine learning for the detection and identification of internet of things (IoT) devices: a survey," \emph{IEEE Internet Things J.}, vol. 7, no. 5, May 2020.

\bibitem{1}
L. U. Khan, W. Saad, Z. Han, E. Hossain and C. S. Hong, "Federated learning for internet of things: recent advances, taxonomy, and open challenges," \emph{IEEE Commun. Surveys Tuts.}, vol. 23, no. 3, Third Quarter 2021.

\bibitem{2}
W. Chen, X. Qiu, T. Cai, H. Dai, Z. Zheng and Y. Zhang, "Deep reinforcement learning for internet of things: a comprehensive survey," \emph{IEEE Commun. Surveys Tuts.}, vol. 23, no. 3, Third Quarter 2021.



\bibitem{4}
G. Jia, G. Han, A. Li and J. Du, "SSL: smart street lamp based on fog computing for smarter cities," \emph{IEEE Trans. Ind. Informat.}, vol. 14, no. 11, Nov. 2018.

\bibitem{6}
A. Farhad, D. Kim and J. Pyun, "R-ARM: retransmission-assisted resource management in LoRaWAN for internet of things," \emph{IEEE Internet Things J.}, 2021.

\bibitem{7}
I. Ilahi, M. Usama, M. O. Farooq, M. U. Janjua and J. Qadir, "LoRaDRL: deep reinforcement learning based adaptive PHY layer transmission parameters selection for LoRaWAN," in \emph{Proc. IEEE LCN}, Nov. 2020.



\bibitem{8}
A. J. Onnumanyi, A. M. Mahfouz and G. P. Hancke, "Cognitive radio in low power wide area network for IoT applications: recent approaches, benefits and challenges,"  \emph{IEEE Trans. Ind. Informat.}, vol.16, no. 12, pp. 7489-7498, Dec. 2020.

\bibitem{9}
A. Li and G. Han, "Full-duplex-based control channel establishment for cognitive internet of things,"  \emph{IEEE Commun. Mag.}, vol.57, no. 3, pp. 70-75, Mar. 2019.
\bibitem{22}
P. Gkotsiopoulos, D. Zorbas and C. Douligeris, "Performance determinants in LoRa Networks: a literature review,"  \emph{IEEE Commun. Surveys Tuts.}, vol.23, no. 3, pp. 1721-1758, third quarter 2021.
\bibitem{23}
J. P. S. Sundaram, W. Du and Z. Zhao, "A survey on LoRa networking: research problems, current solutions, and open issues,"  \emph{IEEE Commun. Surveys Tuts.}, vol.22, no. 1, pp. 371-388, first quarter 2020.
\bibitem{32}
G. Premsankar, B. Ghaddar, M. Slabicki and M. D. Francesco, "Optimal configuration of LoRa networks in smart cities," \emph{IEEE Trans. Ind. Informat.}, vol. 16, no. 12, Dec. 2020.
\bibitem{33}
J. Lim and Y. Han, "Spreading factor allocation for massive connectivity in LoRa systems," \emph{IEEE Commun. Lett.}, vol. 22, no. 4, Apr. 2018.


\bibitem{29}
F. Benkelifa, Z. Qin and J. A. McCann, "User fairness in energy harvesting-based LoRa networks with imperfect SF orthogonality," \emph{IEEE Trans. Commun.}, vol. 69, no. 70, Jul. 2021.
\bibitem{30}
J. Lyu, D. Yu and L. Fu, "Analysis and optimization for large-scale LoRa networks: throughput fairness and scalability," \emph{IEEE Internet Things J.}, 2021.
\bibitem{31}
L. Amichi, M. Kaneko, E. H. Fukuda, N. E. Rachkidy and A. Guitton, "Joint allocation strategies of power and spreading factor with imperfect orthogonality in LoRa networks," \emph{IEEE Trans. Commun.}, vol. 68, no. 6, June. 2020.

\bibitem{25}
B. Su, Z. Qin and Q. Ni, "Energy efficient uplink transmissions in LoRa networks," \emph{IEEE Trans. Commun.}, vol.68, no. 8, pp. 4960-4972, Aug. 2020.
\bibitem{10}
X. Liu, Z. Qin, Y. Gao and J. A. McCann, "Resource allocation in wireless powered IoT networks," \emph{IEEE Internet Things J.}, vol.6, no.3, June 2019.


\bibitem{34}
R. Hamdi, E. Baccour, A. Erbad, M. Qaraqe and M. Hamdi, "LoRa-RL: deep reinforcement learning for resource management in hybrid energy LoRa wireless networks," \emph{IEEE Internet Things J.}, 2021.
\bibitem{35}
H. Fawaz, K. Khawam, S. Lahoud and M. E. Helou, "Cooperation for spreading factor assignment in a multioperator LoRaWAN deployment," \emph{IEEE Internet Things J.}, vol.8, no.7, Apr. 2021.
\bibitem{36}
T. Mai, H. Yao, N. Zhang, W. He, D. Guo and M. Guizani, "Transfer reinforcement learning aided distributed network slicing optimization in industrial IoT," \emph{IEEE Trans. Ind. Informat.}, 2021.

\bibitem{38}
Y. Yu, L. Mroueh, S. Li and M. Terre, "Multi-agent Q-learning algorithm for dynamic power and rate allocation in LoRa networks," \emph{IEEE PIMRC}, 2020.


\bibitem{28}
Y. A. AI-Gumaei, N. Aslam, X. Chen, M. Raza, R. I. Ansari, "Optimising power allocation in LoRaWAN IoT applications," \emph{IEEE Internet Things J.}, 2021.
\bibitem{37}
P. Kumari, H. P. Gupta and T. Dutta, "An incentive mechanism-based stackelberg game for scheduling of LoRa spreading factors," \emph{IEEE Trans. Netw. Service Management}, vol. 17, no. 4, Dec. 2020.

\bibitem{24}
D. Saluja, R. Singh, L. K. Baghel and S. kumar, "Scalability analysis of LoRa network for SNR-based SF allocation scheme," \emph{IEEE Trans. Ind. Informat.}, vol.17, no. 10, pp. 6709-6719, Oct. 2021.
\bibitem{11}
B. Reynders, Q. Wang, P. Tuset-Peiro, X. Vilajosana and S. Pollin, "Improving reliability and scalability of LoRaWANs through lightweight scheduling," \emph{IEEE Internet Things J.}, June 2018.
\bibitem{26}
R. Hamdi, M. Qaraqe and S. Althunibat, "Dynamic spreading factor assignment in LoRa wireless networks," \emph{IEEE ICC}, 2020.
\bibitem{27}
Y. Yu, L. Mroueh, D. Duchemin, C. Goursaud, G. Viver, J. Gorce and M. Terre, "Adaptive multi-channels allocation in LoRa networks," \emph{IEEE Access}, Dec. 2020.


\bibitem{39}
D. Ta, K. Khawam, S. Lahoud, C. Adjih and S. Martin, "LoRa-MAB: toward an intelligent resource allocation approach for LoRaWAN," \emph{IEEE GLOBECOM}, 2019.
\bibitem{12}
A. Azari and C. Cavdar, "Self-organized low-power IoT networks: a distributed learning approach," in \emph{Proc. IEEE GLOBECOM}, Dec. 2018.

\bibitem{42}
A. Li, M. Fujisawa, I. Urabe, R. Kitagawa, S.-J. Kim, and M. Hasegawa, "A lightweight decentralized reinforcement learning based channel selction approach for high-density LoRaWAN," in \emph{Proc. of IEEE DySPAN}, Dec. 2021.
\bibitem{43}
A. Abdelghany, B. Uguen, C. Moy, and D. Lemur, "Decentralized adaptive spectrum learning in wireless IoT networks based on channel quality information," \emph{IEEE Internet Things J.}, DOI: 10.1109/JIOT.2022.3167016, Apr. 2022.
\bibitem{44}
C. Moy, L. Besson, G. Delbarre, and L. Toutain, "Decentralized spectrum learning for radio collision mitigation in ultra-dense IoT networks:LoRaWAN case study and experiments," \emph{Ann. Telecommun.}, vol. 75, pp. 711-727, 2020.

\bibitem{13}
S.-J. Kim and M. Aono, "Amoeba-Inspired Algorithm for Cognitive Medium Access," \emph{IEICE NOLTA}, vol. 5, no. 2, pp. 198-209, 2014.



\bibitem{19}
S.-J. Kim, M. Aono, E. Nameda, "Efficient decision-making by volume-conserving physical object," \emph{NewJ. Phys.}, vol. 17 083023, 2015.

\bibitem{14}
J. Ma, S. Hasegawa, S.-J. Kim and M. Hasegawa, "A reinforcement-leraning-based distributed resource selection algorithm for massive IoT," \emph{Appl. Sci.}, vol. 9, no. 18, pp.3730, Sept. 2019.

\bibitem{15}
D. Yamamoto, H. Furukawa, A. Li, Y. Ito, K. Kato, K. Oshima, S. Hasegawa, Y. Watanabe, Y. Shoji, S.-J. Kim, and M. Hasegawa, "Performance evaluation of reinforcement learning based distributed channel selection algorithm in massive IoT networks," \emph{IEEE ACCESS}, vol. 10, pp. 67870-67882, June 2022.

\bibitem{16}
H. Furukawa, A. Li, Y. Shoji, Y. Watanabe, S.-J. Kim, K. Kato, Y. Andreopoulos and M. Hasegawa, "A channel selection algorithm using reinforcement learning for mobile devices in massive IoT system," in \emph{Proc. IEEE CCNC}, Jan. 2021.


\bibitem{17}
T. Onishi, A. Li, S.-J. Kim and M. Hasegawa, "A reinforcement learning based collision avoidance mechanism to superposed lora signals in distributed massive IoT systems," \emph{IEICE Commun. Exp.}, Mar. 2021.
 
 
 \bibitem{3}
F. Li, D. Yu, H. Yang, J. Yu, H. Karl and X. Cheng, "Multi-armed-bandit-based spectrum scheduling algorithms in wireless networks: a survey," \emph{IEEE Wireless Commun.}, Feb. 2020.

\bibitem{18}
V. K. D. Precup, "Algorithms for the multi-armed bandit problem," \emph{J. Mach. Learn. Res.}, vol. 1, pp.1-48, 2000.

\bibitem{40}
Wi-SUN Alliance. [Online]. Available: https://wi-sun.org/

\bibitem{41}
E. Raeisidehkordi and H. Bakhshi, "Joint fairness and power optimization under user priority in NOMA networks for IoT applications," \emph{IEEE Internet Things J.}, DOI: 10.1109/JIOT.2021.3136758, Dec. 2021.

\end{thebibliography}
\end{document}